\documentclass[11pt]{article}

\usepackage[final]{acl}

\usepackage{times}
\usepackage{latexsym}

\usepackage[T1]{fontenc}

\usepackage[utf8]{inputenc}

\usepackage{microtype}

\usepackage{inconsolata}

\usepackage{graphicx}

\usepackage{hyperref}
\usepackage{longtable}
\usepackage{multirow}
\usepackage{geometry}
\usepackage{array}
\usepackage[table]{xcolor}
\usepackage{subcaption}

\usepackage{xcolor}

\def\cca#1{%
    \pgfmathsetmacro\calc{(#1-0.1)*100/(0.9-0.1)}%
    \edef\clrmacro{\noexpand\cellcolor{black!\calc}}%
    \clrmacro%
    \ifdim \calc pt>50pt\color{white}\fi{#1}%
}

\usepackage{float}
\usepackage{newfloat}
\usepackage{listings}
\floatstyle{ruled}
\newfloat{listing}{tb}{lst}{}
\floatname{listing}{Listing}
\title{Does Local News Stay Local?: Online Content Shifts in \\Sinclair-Acquired Stations}

\author{Miriam Wanner$^*$,\quad Sophia Hager$^*$,\quad Anjalie Field \\ \texttt{\{mwanner5, shager2, anjalief\}@jhu.edu}\\ Johns Hopkins University}

\begin{document}
\maketitle

{
\def\thefootnote{*}\footnotetext{Equal contribution.}
}

\begin{abstract}
Local news stations are often considered to be reliable sources of non-politicized information, particularly local concerns that residents care about. 
The Sinclair Broadcast group is a broadcasting company that has acquired many local news stations in the last decade. We investigate the effects of local news stations being acquired by Sinclair: how does coverage change?
We analyze YouTube content put out by local news stations through topic modeling, log-odds ratios, and word embedding analyses to investigate changes after being acquired by Sinclair. We find evidence that local news stations report more frequently on national news at the expense of local topics, and that their coverage of polarizing national topics increases. These findings associate acquisition by Sinclair with increasing polarization and nationalization of news content, which in-turn risks increasing political polarization of local news viewers.
\end{abstract}

\section{Introduction}

Historically, local news outlets have played a vital role in the news ecosystem for many Americans by providing information that is community-focused with less perceived partisanship than national outlets. Viewers find local news topics like weather, local crime, and traffic reports important to know about for daily life \cite{pew_local_news_demographic}. American adults also tend to view local news positively regardless of political affiliation, whereas there are stark political divides in opinions about national news \cite{pew_local_news_attitude}. Furthermore, local news consumption has been associated with greater knowledge of local election candidates and increased likelihood of voting for candidates from different political parties for state governor and U.S. president rather than solely along party lines \citep{MOSKOWITZ_2021}.

The Sinclair Broadcast Group, one of the largest broadcasting companies in the United States, owning or operating 185 stations,\footnote{\url{https://sbgi.net}} has acquired a number of local news stations, with purchases primarily concentrated around 2000, 2012-14, and 2016-17. 
These acquisitions and subsequent observations of news coverage have raised concerns around ways Sinclair is influencing local news.
Outside reporters have exposed Sinclair for requiring stations to run specific video segments or to deliver the same scripted speech, and they accused the company of right-wing bias.\footnote{\url{https://www.nytimes.com/2018/04/02/business/media/sinclair-news-anchors-script.html?searchResultPosition=16}}  
Researchers have similarly identified conservative bias \citep{tryon-sinclair}, and demonstrated that Sinclair stations produce more stories with dramatic elements, commentary, and partisan sources than non-Sinclair stations \citep{hedding2019sinclair}.
Concerningly, there is also evidence that Sinclair takeovers actually influenced viewers perceptions of politicians \citep{levendusky2022local}.




Given the importance of local news and the growing Sinclair influence, we investigate the effect that acquisition by Sinclair has on the content of local news stations. We compare content in news stations before and after Sinclair purchases, and we further draw comparisons with national news outlets. We focus on two levels of analysis:

\begin{enumerate}
    \item How does overall news differ after purchase?
    \item How does coverage of politicized topics differ after purchase?
\end{enumerate}

\begin{table*}[ht]
    \begin{center}
    \small
    \begin{tabular}{cp{1.2in}cccc}
        TV Channel & City & Purchased & Affiliation & Youtube & \#Videos \\
        \hline \hline

        \rowcolor{gray!40} \multicolumn{6}{c}{TV Channels Purchased by Sinclair} \\
        \hline
        WSBT-TV & South Bend, IN & 02/12/16 & Fox & @wsbttv & 10624 \\
        
        \rowcolor{gray!10} KECI/KCFW/KTVM & Missoula, MT; Kalispell, MT; Butte, MT & 09/01/17 & NBC & @NBCMontana & 3314 \\
        
        WCTI-TV & Greenville, NC; New Bern, NC; Morehead City, NC & 09/01/17& ABC & @WCTI& 3320\\
        \rowcolor{gray!10} WCYB&Bristol, VA; Greenville, TN; Johnson City, TN; Kingsport, TN & 09/01/17& NBC/The CW &@wcyb5 & 4620\\
        WLUK-TV&Green Bay, WI &12/19/14 & Fox & @Fox11online& 19774\\        
        \rowcolor{gray!10} WJAR& Providence, RI; New Bedford, MA& 12/19/14 & NBC &@NBC10WJAR & 5044\\
        WGXA & Macon, GA & 09/03/14 & Fox/ABC & @WGXA & 2063\\
        \rowcolor{gray!10} WJLA-TV& Washington, DC& 08/01/14 & ABC & @7NewsDC & 19425\\
        \hline \hline
        \rowcolor{gray!40} \multicolumn{6}{c}{Left- and Right-Wing TV Channels for Comparison} \\
        \hline
        CNN & & & & @CNN & 27560 \\
        \rowcolor{gray!10}Fox & & & & @FoxNews & 29986 \\
    \end{tabular}
    \caption{Summary data statistics. We collected transcripts from 8 geographically diverse local news YouTube channels that were purchased by Sinclair, as well as transcriptions from YouTube channels for two national outlets.}
    \label{tab:stats}
    \end{center}
\end{table*}

While a small amount of prior work has compared broadcasts in Sinclair-owned and non-Sinclair stations \citep{martin2019local,hedding2019sinclair} or news station websites \citep{blankenship2021corporate}, Americans are increasingly viewing digital local news, rather than obtaining it through broadcast television or radio \cite{pew_local_news_attitude}.
Thus, we focus on a novel data source: news station YouTube channels, allowing us to uniquely examine the content that news stations choose to highlight on social media and if it reflects trends in broadcast data. Our dataset contains data from eight stations over sixteen years of publishing videos. This construction allows us to examine differences in coverage within the same station before and after acquisition as well as between the larger group of Sinclair-affiliated and non-affiliated stations at any particular point in time. We further include two national news outlets (Fox News and CNN), enabling direct comparisons of Sinclair-owned local news and national news.


We use a combination of corpus analysis methods to examine overall shifts in content and target politicized topics, including comparisons of word choice \cite{Monroe_Colaresi_Quinn_2017}, topic modeling with covariates \citep{roberts2013structural,roberts2019stm}, and word embeddings analyses \cite{mikolov2013efficient,garg2018word}.
We find compelling evidence that after purchase, news channels move from covering mostly local topics to politicized national topics.
Overall our work offers insight into the content changes associated with Sinclair purchases, thus contributing understanding of how the purchases may influence viewers and highlighting the urgent decline of community-focused news.

\begin{figure*}[ht]
    \centering
    \includegraphics[width=0.99\textwidth]{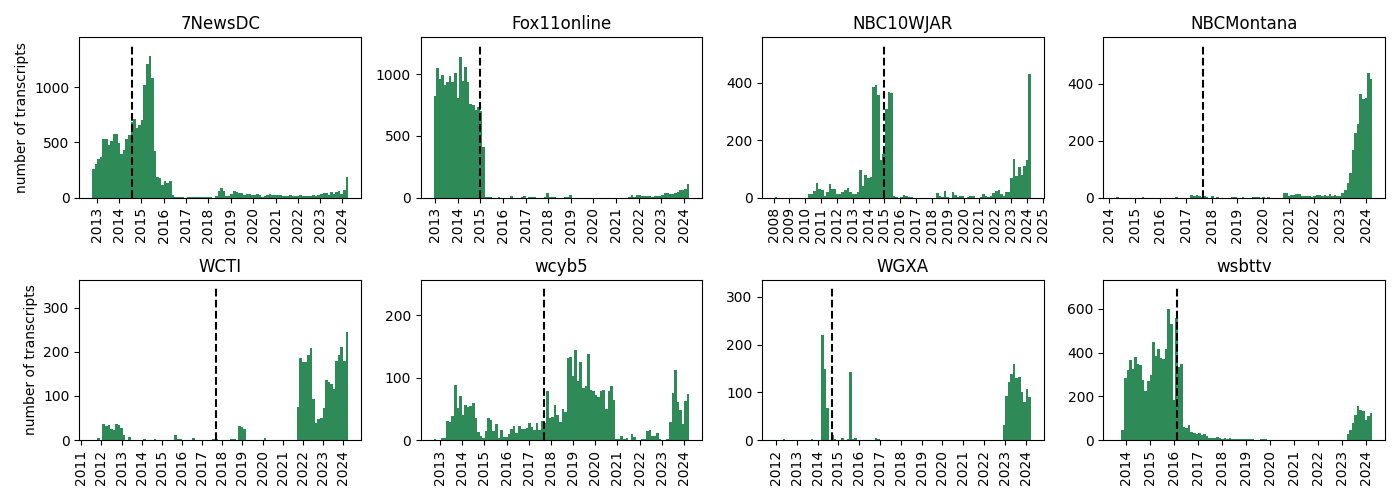}
    \caption{The distribution of the data by year. Vertical lines denote the date that the station was purchased by Sinclair.}
    \label{fig:data}
\end{figure*}

\section{Related Work}

\paragraph{Sinclair Broadcast Group} While the Sinclair Broadcast Group's takeover of local news stations attracted public interest and journalism, analysis of content differences in news coverage have been limited to a few prior studies.
\citet{martin2019local} use topic modeling and comparisons of phrases with U.S. Congressional Record \citep{gentzkow2010drives} to show Sinclair ownership is associated with a drastic increase in national political coverage over local political coverage and a right-wing shift in ideology.
\citet{blankenship2021corporate} similarly find decreased coverage of local news in their analysis of locations mentioned in news stories on 
 Sinclair-owned station websites, though decreasing local news coverage predates Sinclair ownership and coincides with reposted content.
In a content-focused analysis,  \citet{hedding2019sinclair} find that  Sinclair stations produce more stories with dramatic elements, commentary, and partisan sources.
None of these studies focus on YouTube data or conduct an in-depth language analysis focusing on politicized topics.


A related line of work has focused on the effects of Sinclair purchases on viewers, without examining content changes in news coverage. \citet{miho2018} examines the effect of Sinclair ownership on election results. Using event study methodology, they find a 2.5\%-point increase in the Republican vote of the 2008/2012 elections, with double the increase in the 2016/2020 elections as a result from exposure to Sinclair content starting in 2004.
\citet{levendusky2022local} use statistical methods to find that living in an area with a Sinclair-owned TV station reduces viewers' approval of President Obama. They find lower approval during his time in office, and additional evidence that viewers are then less likely to vote for the presidential Democratic nominee. These findings that Sinclair purchases are associated with observable changes in preferences of viewers motivate our investigation into understanding the language and content changes that may be driving them.

\paragraph{U.S. Local news}  Concerning local news more generally, there has been a documented decline in local news organizations, leading to growing ``news deserts'': areas without consistent news coverage \citep{abernathy2016rise,abernathy2018expanding}.
There is evidence that declining local news is contributing to political polarization. Local news consumption is associated with decreased voting exclusively along party lines  \citep{MOSKOWITZ_2021}, while increased coverage of local content is associated with lower feelings of political divide \citep{darr2021home}. These factors add further motivation to understanding content and language changes in Sinclair-owned stations.
In-depth text analyses of local news have focused coverage of the COVID-19 pandemic \citep{horne2022nela} and the creation of datasets for further investigation \citep{joseph2022local}. These studies are less related to our work but generally validate interest in understanding local news coverage.

\paragraph{U.S. Media polarization} 
Analyses of a variety of text data, including social media posts \citep{demszky-etal-2019-analyzing} and political speeches \citep{card2022computational} has uncovered evidence of increasing polarization in the U.S. Despite early evidence of media slant in news articles \citep{gentzkow2010drives} and many anecdotes about media bias, fewer quantitative analyses have focused on evidence of polarization from video footage.
The Stanford Cable TV NewsAnalyzer \citep{cabletvnews} offers an extensive dataset for examining content in three U.S. cable news networks (CNN, Fox, and MSNBC).
\citet{ding2023sem-polar} use this data to evaluate the semantic polarization in online public discourse, finding that CNN and Fox News cover similar topics, however with varying, distinct contexts, reflecting the polarization between the political leaning of these news stations. They also show that polarization sharply increases around 2016, with its highest peak in 2020, aligning with the death of George Floyd and following Black Lives Matter demonstrations.
These findings motivate our use of CNN and Fox News as comparisons datasets in our analysis of local news.

\section{Dataset}

\paragraph{Collection}
We construct a new dataset consisting of
 automated closed captions from YouTube channels of news stations.
 First, we identified news stations that were acquired Sinclair by starting from an initial list\footnote{\url{https://en.wikipedia.org/wiki/List_of_stations_owned_or_operated_by_Sinclair_Broadcast_Group}} and retaining only stations that (1) have a YouTube channel and (2) began posting videos before they were purchased by Sinclair. We identified 8 local news stations for analysis, as well as Fox News and CNN for comparison. 
 For each station, we download YouTube closed captions for all videos on the channel.
  We preprocessed this data by converting all transcripts into lowercase and removing common non-speech tokens or utterances unlikely to provide meaningful signal.\footnote{``$>$,'' ``[music],'' ``[applause],'' ``uh''}

\autoref{tab:stats} reports the full list of stations, their purchase date, and the number of identified videos. Four stations were purchased in 2014, one was purchased in 2016, and three were purchased in 2017. The stations are geographically diverse, reflecting various cities in the east and central U.S. While there is variance in the amount of data from each each station, our data contains at least 2,000 videos for each station.
In \autoref{fig:data}, we further show how the transcripts for each local news station are distributed over time, relative to the data of Sinclair purchase. For some stations (e.g., 7NewsDC, NBC10WJAR) there is a concentration of data just before and just after purchase. For other stations (wcyb5) the data is more dispersed over time. 

\begin{table*}[ht]
    \begin{center}
    \small
    \begin{tabular}{p{3.2in}p{3.2in}}
        Non-Sinclair & Sinclair \\
        \hline \hline

        \rowcolor{gray!40} \multicolumn{2}{c}{2014} \\
        \hline
        so, little, it's, it, okay, green, really, bit, bay, nice,yeah, then,  we're, going, great, just, can, snow, fun, kind & 7, he, virginia, \textbf{president}, washington, police, matthew, who, his, was, jury, that, wilson, live, williams, united, robert, said, quarterback, thank\\
        
        \hline \hline
        \rowcolor{gray!40} \multicolumn{2}{c}{2015} \\
        \hline
        \rowcolor{gray!10}south, 22, patrick, jennifer, st., kelly, football, james,  desk, st, accurate, season,  first, watching, downtown,play, says, at, year & you, i, that, 7, okay, \textbf{government}, washington, what, we, \textbf{trump}, going,  island, \textbf{president}, virginia, think, \textbf{federal}, let's, sam, bay, of\\

        \hline \hline

        \rowcolor{gray!40} \multicolumn{2}{c}{2016} \\
        \hline
         school, snow, tonight, ice, animals, girls, cold, coach, st., submit, chevy, morning, sale, home, kids, church, temperatures, at, 22, lead & that, \textbf{trump}, think, \textbf{government}, \textbf{federal}, \textbf{president}, of, \textbf{republican}, i, \textbf{sanders}, states, going, security, \textbf{campaign}, sort, there's, is, terms, \textbf{voters}, \textbf{political} \\
    \end{tabular}
    \caption{Fightin' words results broken down by year. Words that are likely relevant to broader political concerns in the United States have been bolded. Sinclair-purchased stations tend to have more of these words, while stations that are not owned by Sinclair tend to discuss more local concerns.}
    \label{tab:fightingwords}
    \end{center}
\end{table*}

\section{RQ1: How does overall news differ after purchase?}

We first use exploratory text analysis methods to broadly examine how news coverage differs before and after Sinclair purchase, as well as in comparison to the two national outlets.

\subsection{Methods}\label{sec:methods1}

We use two primary methods for examining overall news coverage. First, we examine words that are overrepresented in data before purchase as compared to after purchased using log-odds ratio with a Dirichlet prior (method referred to as ``Fightin' Words''; from \citet{Monroe_Colaresi_Quinn_2017}). We preprocess the transcripts by filtering out words that do not appear at least ten times in transcripts from every station.
A potential confounder is that news changes over time, and our data tends to have more Sinclair-owned stations as time goes on. To mitigate this, we stratify our data by year and only compare log-odds over the years where we have a reasonable amount of paired data (2014, 2015, and 2016). The year with the most paired data, 2014, contains 19,936 videos, 16,640 from stations before they are purchased, and 3,296 from already purchased stations. Paired data decreases in the following couple years, as many stations are purchased during this time period. Purchased stations become more prevalent in 2015 data with 8,641 videos from after purchase, and 4,281 before. This imbalance is more pronounced in 2016 with 696 videos before purchase, and 1,788 after.

Second, we use topic models to examine coverage changes in clusters of co-occurring words, rather than just individual words. We specifically use the Structured Topic Model (STM) \citep{roberts2013structural,roberts2019stm}, which is an extension of the popular Latent Dirichlet Allocation (LDA) \citep{lda}, that flexibly incorporates document metadata as covariantes. We chose this model for this property, as well as based on evidence that classical LDA-style models achieve better stability and alignment with human annotations than more recent neural alternatives \citep{hoyle-etal-2022-neural}.

We train five STM models on the transcripts of Sinclair-owned stations, non-Sinclair affiliated stations, Fox, and CNN, in order to determine how the topics discussed by these stations change.\footnote{In appendix section \ref{sec:paired-analysis} we further conduct a controlled pairwise comparison between two Sinclair-purchased stations and two non-purchased stations to isolate the effect of purchase.} For the first four models, we use news-affiliation (which includes four options: Before Sinclair purchase, After Sinclair purchase, CNN, Fox), and date as covariates. We use the convenience function to select a flexible b-spline basis for the date covariate. These four models only differ in the subset of data used. First, we use data from all dates collected. Models 2-4 use data only from 2014, 2015, and 2016, respectively.
With these models, we evaluate the topic prevalence, and plot the difference of prevalence along two axes: (1) Before Purchase - After Purchase, and (2) CNN - Fox, in order to highlight topic relationships between Sinclair owned stations and political leaning. We remove the 1\% most sparse and common words, and use 30 topics for all models, which we found to have the most coherent topics.

A shortcoming of the STMs introduced thus far is that they assume a fixed vocabulary distribution within topics. If we are interested in comparing the how discourse differs for the same topic, we need to allow these distributions to vary within topics. To study this, we train a 5th model across all the data where we let the influence of Sinclair-affiliated versus non-Sinclair-affiliated (excluding CNN and Fox) be a topical content covariate. We can then then look at the difference in prevalence of words between the content covariate for a given topic. We use the same data filtering and number of topics as in the previous models.

\subsection{Results}

The Fightin' Words analysis is shown in \autoref{tab:fightingwords}. Stations that have been purchased by Sinclair are more likely to use words relevant to national politics, rather than local concerns, using words such as ``president,'' ``government,'' or ``federal.'' Non-Sinclair 
(pre-purchase) stations were more likely to discuss events that were relevant to local viewers, such as weather, sports, and school, using words like ``snow,'' ``downtown,'' or ``school.'' Some of this may be due to the timing of the purchase coinciding with events occurring in the year (particularly 2016, as an election year). Notably, however, Sinclair-owned stations in 2015 were more likely to discuss the national election than non-Sinclair stations in the election year of 2016, demonstrating that the timing of purchase by Sinclair cannot fully explain the shift to discussing national topics.

\begin{figure*}[htbp]
    \centering
    \begin{subfigure}[b]{0.24\linewidth}
        \centering
        \includegraphics[width=\linewidth]{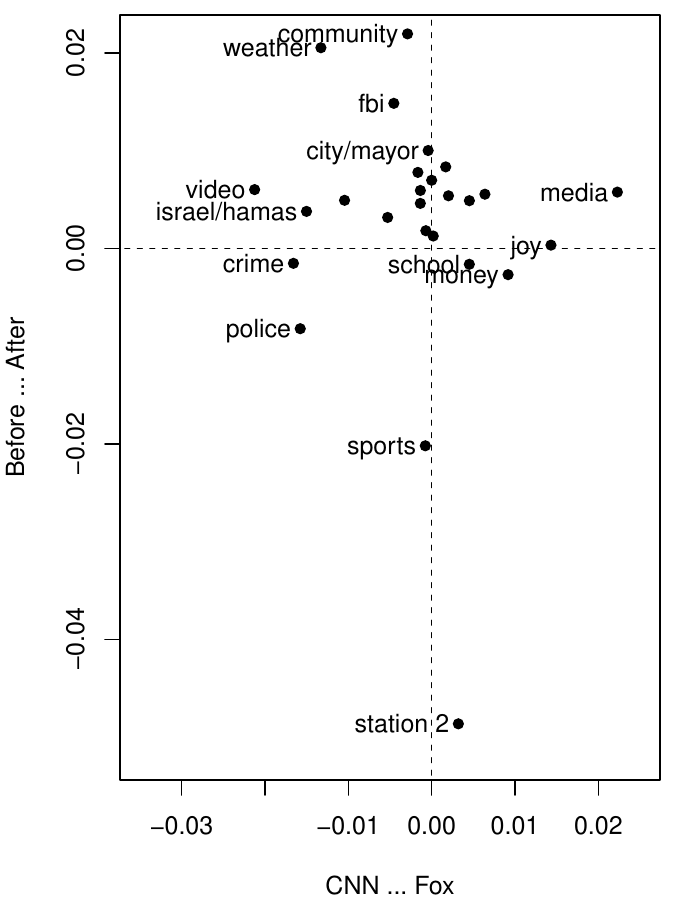}
        \caption{All years}
        \label{fig:stm-all-main}
    \end{subfigure}
    \hfill
    \begin{subfigure}[b]{0.24\linewidth}
        \centering
        \includegraphics[width=\linewidth]{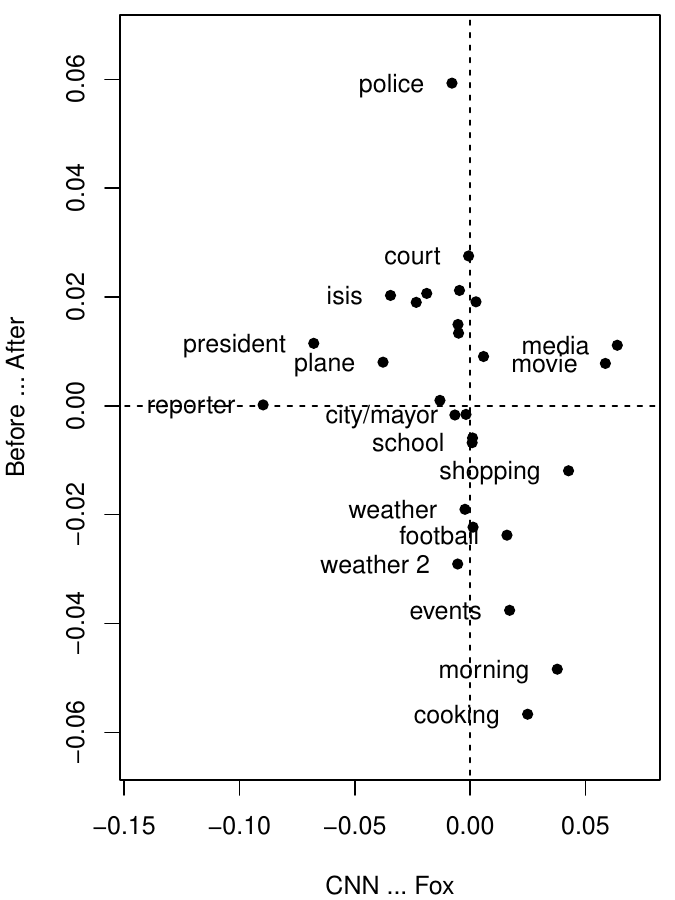}
        \caption{2014}
        \label{fig:stm-2014-main}
    \end{subfigure}
    \hfill
    \begin{subfigure}[b]{0.24\linewidth}
        \centering
        \includegraphics[width=\linewidth]{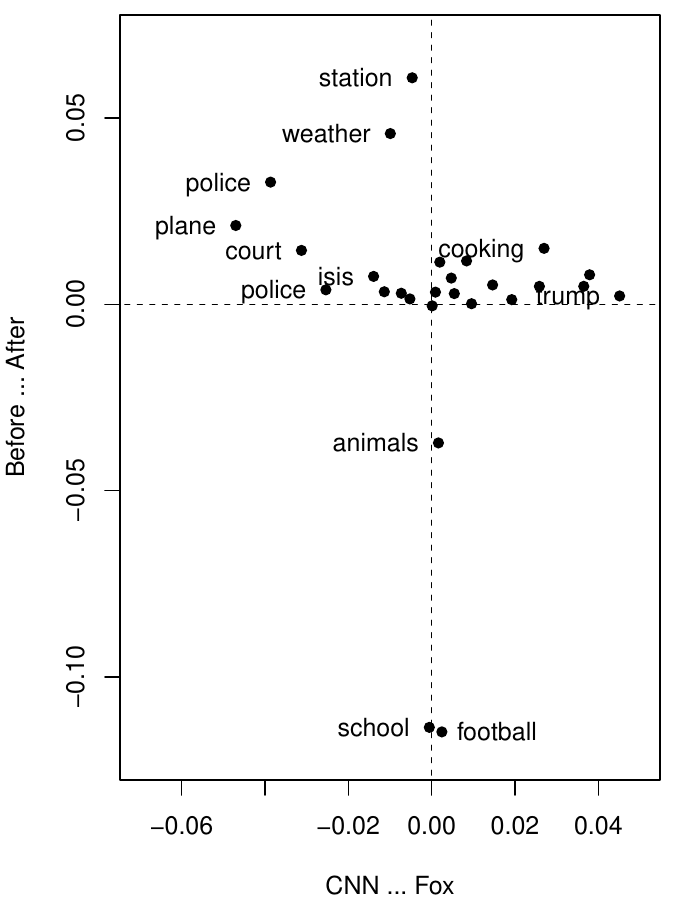}
        \caption{2015}
        \label{fig:stm-2015-main}
    \end{subfigure}
    \hfill
    \begin{subfigure}[b]{0.24\linewidth}
        \centering
        \includegraphics[width=\linewidth]{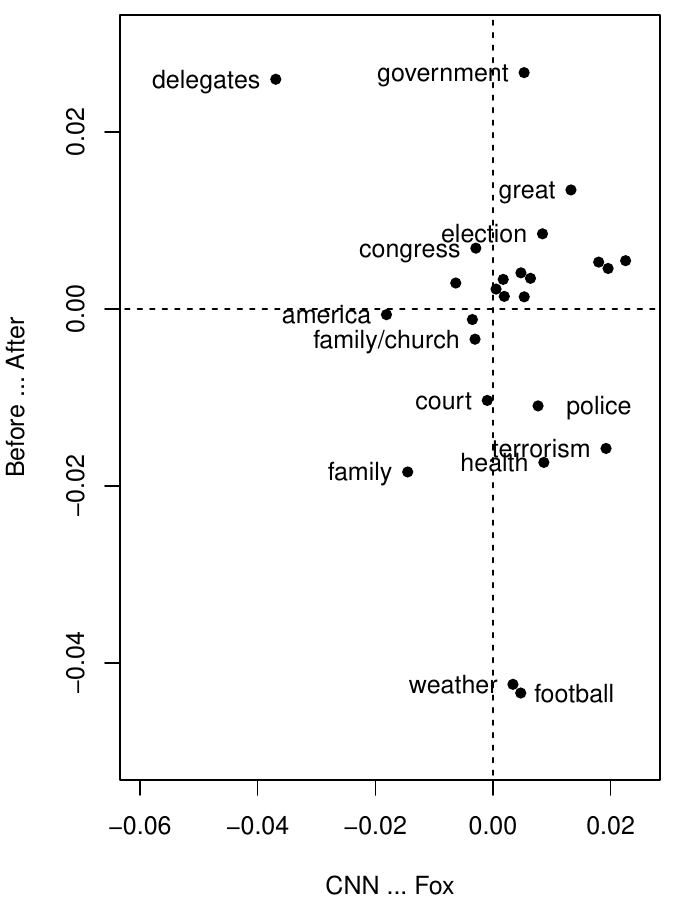}
        \caption{2016}
        \label{fig:stm-2016-main}
    \end{subfigure}
    
    \caption{Results for STM on all, 2014, 2015, and 2016 data subsets. Change in topic proportion shifting from CNN to Fox on the x-axis and before Sinclair purchase to after Sinclair purchase on the y-axis. Coverage becomes more national and political in stations purchased by Sinclair, with stations before purchase discussing local topics.}
    \label{fig:stm-main-2x2}
\end{figure*}





\paragraph{Topic Models}
STM results for selected topics are shown in Figures \ref{fig:stm-all-main}-\ref{fig:stm-2016-main}, where we manually assign representative names for each topic. In the appendix, we report  results for all topics (Figures \ref{fig:stm-all-appendix}-\ref{fig:stm-2016-appendix}) and complete lists of probable words for each topic (Tables \ref{tab:topics-all}-\ref{tab:topics2016}). These plots display change in topic proportion between CNN and Fox data, and before and after Sinclair purchase. 

Coverage by stations acquired by Sinclair becomes more politicized compared to their reporting before purchase. Topics most widespread on stations not Sinclair affiliated include local topics about the local community, including school (\ref{fig:stm-2014-main}, \ref{fig:stm-all-main}), family (\ref{fig:stm-2016-main}), and local events (\ref{fig:stm-2014-main}). Discourse on the local environment is also prevalent before purchase with topics including weather (\ref{fig:stm-2014-main}, \ref{fig:stm-2016-main}), animals (\ref{fig:stm-2015-main}), and health (\ref{fig:stm-2016-main}). Topics of local interest include football (\ref{fig:stm-2014-main}, \ref{fig:stm-2015-main}, \ref{fig:stm-2016-main}), other sports (\ref{fig:stm-2014-main}, \ref{fig:stm-all-main}), and cooking (\ref{fig:stm-2014-main}), and are also mostly featured before purchase. Finally, we see topics around local station small talk, including morning conversation (\ref{fig:stm-2014-main}) and station specific language (\ref{fig:stm-all-main}).
After stations are acquired, topics predominantly covered shift, with a more political and national focus. Discourse on presidential candidates, including Clinton and Trump are covered at a higher rate on Sinclair-affiliated stations (\ref{fig:stm-2014-appendix}), in addition to other national topics like the FBI (\ref{fig:stm-all-main}). In additional to national news, these Sinclair-owned stations also discuss US-relevant international news, including ISIS (\ref{fig:stm-2014-main}) and terrorism (\ref{fig:stm-2015-appendix}). These Sinclair purchased stations still cover local news including family/church (\ref{fig:stm-2014-appendix}, \ref{fig:stm-all-appendix}), community, city/mayor information, and football (\ref{fig:stm-all-appendix}), indicating that although these stations appear to increasingly report on politicized topics, there is still some local news coverage.
A few topics are widely covered and do not tend to align consistently with either before/after purchase. Police and crime are often reported on, and are prevalent both before (\ref{fig:stm-2016-main}) and after (\ref{fig:stm-2014-main}) coverage, and are particularly aligned with CNN (\ref{fig:stm-all-main}, \ref{fig:stm-2015-main}). This trend is also reflected in proportion of weather discourse, prevalent both before (\ref{fig:stm-2014-appendix}) and after (\ref{fig:stm-all-main}) Sinclair purchased stations.

Overall, the difference in topic proportion is less pronounced on the CNN/Fox axis. In fact, topics reported on pre-purchased stations tend to be not clearly aligned with either CNN or Fox, indicating these stations are reporting equally, and possibly very little, on these topics. However, there exists more variation on the CNN/Fox axis in station coverage after Sinclair purchase, and a few are more topics predominantly covered by Fox or CNN. 
For instance, the conflicts between Israel and Hamas appeared to be reported on more by CNN (\ref{fig:stm-all-main}), in addition to other politicized topics like the president (\ref{fig:stm-2014-main}) and delegates in congress (\ref{fig:stm-2016-main}). In 2014, Fox reports more on media and movies (\ref{fig:stm-2014-main}), which are not inherently political topics. This balance shifts, when in 2015 Trump is a more prevalent topic reported by Fox (\ref{fig:stm-2015-main}). This topic shift is further seen in 2016, with topics like ISIS, Cuba, immigration, and presidential candidates being covered by Fox more (\ref{fig:stm-2016-appendix}).

Not only can we observe these interactions and changes of topic proportion on the two axes, but we can also see topics shift and appear over time. The Ebola outbreak of 2014 emerges as a topic (\ref{fig:stm-2014-appendix}) in the 2014 STM (\ref{fig:stm-2014-main}), mostly covered by stations purchased by Sinclair. Leading up to the election, presidential candidates, debates, debate topics (e.g. immigration, china) emerge as topics in 2015 and 2016, similarly covered by Sinclair-owned stations (\ref{fig:stm-2015-main}, \ref{fig:stm-2016-main}). 

From our STM with non- and Sinclair owned as a topical context covariate, we can observe differences in the language used to discuss certain topics. The results can be found in tables \ref{tab:content-topics}-\ref{tab:content-topics-beforeafter}, and figure \ref{fig:content-stm-words} in the appendix. Discourse on certain topics differs between stations. 
Coverage of disasters and violence are covered by before-purchase stations in discussions of healthcare, and after-purchase stations in discussions of topics such as war and terrorism (\ref{fig:content-topic11}).
Similarly, discussions of crime appear typical in stations not affiliated with Sinclair, but purchased stations include discussions of protest and protesters (\ref{tab:content-topics-beforeafter}). 
A topic on general government infrastructure (\ref{fig:content-topic26}) uses language about money (taxes, dollars, etc.) in pre-purchase coverage, and is more national after purchase, using words like pentagon and intelligence. 
Discourse on legislature differs, with non-Sinclair owned stations talking about local politics including the mayor or local candidates, and Sinclair stations using words like immigration and shutdown, seemingly more national (\ref{fig:content-topic27}). 
Discourse on photos and cameras shifts towards a theme of surveillance after purchase, with pre-purchase stations using language like pictures (\ref{fig:content-topic1}). 

\begin{table*}[ht]
    \begin{center}
    \small
    \begin{tabular}{p{0.9in}|p{0.9in}||p{0.9in}|p{0.9in}||p{0.9in}|p{0.9in}}
        
        \rowcolor{gray!40} \multicolumn{2}{c}{\textbf{white}} &\multicolumn{2}{c}{\textbf{black}} &\multicolumn{2}{c}{\textbf{bias}}\\
        \hline
        \rowcolor{gray!25}Before & After &Before & After &Before & After  \\
        \hline \hline
        black       & supremacist    & white      & africanamerican  & chiffonade  & implicit       \\
\rowcolor{gray!10}red         & supremacy      & tan        & racial           & basil       & racial         \\
yellow      & secret         & hoodie     & africanamericans & greens      & perpetuate     \\
\rowcolor{gray!10}blue        & house          & sweatshirt & color            & noir        & racist         \\
velvet      & presidents     & wearing    & blacks           & vinaigrette & discrimination \\
\rowcolor{gray!10}burgundy    & clancy         & dark       & colored          & italian     & racially       \\
roses       & obamas         & colored    & racism           & mince       & quote          \\
\rowcolor{gray!10}orange      & obama          & stripes    & brown            & riesling    & shameful       \\
colored     & pierson        & bandana    & african          & baguette    & prejudice      \\
\rowcolor{gray!10}chardonnay  & omar           & yellow     & movement         & seedless    & language       \\
        \hline
        \rowcolor{gray!40} \multicolumn{2}{c}{\textbf{climate}} &\multicolumn{2}{c}{\textbf{equality} }&\multicolumn{2}{c}{\textbf{abortion}}\\
        \hline
        \rowcolor{gray!25}Before & After &Before & After &Before & After  \\
        \hline \hline
        growth      & fuels          & rights     & freedom          & abortions   & abortions      \\
\rowcolor{gray!10}industry    & emissions      & gay        & rights           & parenthood  & roe            \\
uncertainty & global         & marriage   & prolife          & privileges  & reproductive   \\
\rowcolor{gray!10}algae       & fossil         & religious  & dignity          & admitting   & wade           \\
economy     & everglades     & democracy  & equal            & ultrasound  & prolife        \\
\rowcolor{gray!10}regionally  & sustainability & moral      & freedoms         & gyn         & prochoice      \\
ratings     & pollution      & civil      & unborn           & prolife     & overturning    \\
\rowcolor{gray!10}potential   & droughts       & marriages  & racism           & clinic      & marriage       \\
consumption & environmental  & supreme    & democracy        & clafer      & affirming      \\
\rowcolor{gray!10}economic    & impacts        & samesex    & lgbt             & pregnancies & incest \\

    \end{tabular}
    \caption{Ten nearest neighbors to the query word (bold) for stations that were not owned by Sinclair (Before) and Sinclair-owned stations (After). Sinclair-owned stations are more likely to have nearest neighbors that are politically charged.}
    \label{tab:n_neighbors}
    \end{center}
\end{table*}




\section{RQ2: How does coverage of politicized topics differ after purchase?}
\subsection{Methods}\label{sec:methods2}

In order to examine shifts in coverage of politicized topics, we train word embedding models and examine properties of embeddings for selected keywords, using similar methodology as prior word embedding analyses \citep{garg2018word,rodriguez-embeddings}.
We train  separate Word2Vec models \citep{mikolov2013efficient} for all transcripts of stations before Sinclair purchase and all transcripts after purchase. We additionally train embedding models for data from CNN and Fox News, for left- and right-wing news station comparison.\footnote{For all embedding models we use the following hyperparameters: \texttt{window=50}, \texttt{min\_count=10}, \texttt{seed=42}, \texttt{workers=16}, \texttt{vector\_size=100}. Prior work has shown exact parameter settings have little impact on analysis results \citep{rodriguez-embeddings,joseph-morgan-2020-word}.}

We curate a set of keywords related to politicized issues for analysis, following \citet{rodriguez-embeddings}. We start with their set of words pertaining to policy issues that are debated by political parties and motivate voting: ``immigration,'' ``abortion,'' ``welfare,'' ``taxes.'' We add words relating to policy issues not covered in their original set, including words related to racial bias (``racism,'' ``bias,'' ``black,'' ``white''), ``climate,'' ``police,'' ``military,'' and ``guns.''
We further include words that \citet{rodriguez-embeddings} curate as expected to solicit different response in different people: ``democracy,'' ``freedom,'' ``equality,'' ``justice,'' ``republican,'' and ``democrat,'' though they are less relevant to our focus on politicized topics. We identify the 10 nearest neighbors for each keyword in the before-purchase and after-purchase embedding models using cosine similarity. 

\paragraph{Embedding Similarity} We conduct an analysis of whether increased politicization can be noted in our learned word embeddings when put in context of national news stations. We examine how similar word embeddings are for the seed words described in \autoref{sec:methods2} in comparison to two national news sources, Fox News and CNN. We choose Fox News and CNN in particular as they are considered to be politically polarized \citep{ding2023sem-polar} and thus are likely to be talking about politically polarizing issues. In this experiment, we train embedding models on data from stations before purchase and after purchase specifically between the years 2014-2016 and evaluate similarity between embeddings from before/after purchase trained models with models trained on Fox News/CNN transcripts from the same time period. We query these models with the mentioned seed words, and align the embedding spaces and their vocabularies using the Procrustes transformation. We then calculate cosine similarity between embedding vectors for the same word, to determine whether local news outlets tend to be more likely to discuss these words in similar contexts to the polarized national news outlets.

\subsection{Results}

\paragraph{Nearest Neighbor Analysis}
We show the six most interpretable nearest neighbor results in \autoref{tab:n_neighbors}, with the remaining twelve less-interpretable results presented in \autoref{tab:more-neighbors}. We find that the embedding model trained on transcripts from stations after being purchased by Sinclair tends to have nearest neighbors to our query words that are more overtly politically charged. The first three words we show,``white,'' ``black,'' and ``bias,'' demonstrate the clearest movement towards polarizing rhetoric. Before purchase, ``white'' and ``black'' are generally associated with other colors and patterns (e.g.``red,'' ``stripes'') or items that might be that color (e.g.``chardonnay,'' ``roses''). The embedding model trained on data after Sinclair acquisition associates black mostly with words pertaining to race (``africanamerican,'' ``racism''), as well as ``movement,'' likely relevant to protest movements such as Black Lives Matter. While ``supremacist''/``supremacy'' are the nearest neighbors to ``white,'' indicating increased use of white as a racial descriptor, many of the nearest neighbors appear to be relevant to the presidency (``house,'' ``obama'') indicating an increased discussion of White House policy and national political news. We note a similar result with the query word ``bias.'' The nearest neighbors before acquisition associate ``bias'' predominantly with cooking, and with cutting in particular (``chiffonade,'' ``mince''), likely due to the phrase ``cutting on the bias'' being frequently used as an instruction in cooking videos. The model trained on data post-acquisition associates ``bias'' with words more evocative of societal bias (``implicit,''  ``prejudice''). 

\autoref{tab:n_neighbors} also displays results for ``climate,'' ``equality,'' and ``abortion,'' which demonstrate some signs of increased politicization, but may be influenced by the confounding factor of time. For instance, the post-acquisition embedding model was more likely to associate ``climate'' with words referencing climate change (``emissions,'' ``pollution''), while the pre-acquisition model generally referenced other topics (``growth,'' ``economy,''). This may indicate increased discussion of global warming, but may also be influenced by increased discussion of climate change in recent years.
The pre-acquisition model associates ``equality'' with an assortment of words which suggest discussions of Obergefell v. Hodges (2015)\footnote{\url{https://www.justice.gov/sites/default/files/crt/legacy/2015/06/26/obergefellhodgesopinion.pdf}, a Supreme Court decision which legalized gay marriage in the United States},  such as ``marriages'' and ``supreme.'' After, some neighbors are relevant to reproductive justice (``prolife,'' ``unborn''), possibly due to conversations about Dobbs v. Jackson Women's Health Organization (2022)\footnote{\url{https://www.supremecourt.gov/opinions/21pdf/19-1392_6j37.pdf}, a Supreme Court decision which overturned Roe v. Wade by asserting there was no constitutional right to abortion}. While this may indicate increased discussion of reproductive justice, it also may demonstrate confounds of this data. This case may also explain the shift in nearest neighbors to ``abortion''; the pre-acquisition model associates ``abortion'' with more healthcare-related words (e.g.``admitting''+``privileges,'' ``ultrasound'') while the post-acquisition embedding model associates it more with politicized rhetoric around abortions (e.g.``prolife,'' ``prochoice'').

In summary, the analysis of the differences in embedding models trained on these transcripts offers clear evidence that stations owned by Sinclair  discuss polarizing political issues more than ones that have not been purchased. While we cannot necessarily attribute Sinclair purchase as the sole cause of this coverage shift, it nevertheless indicates increasing politicization of local news.


\begin{figure}[ht]
    \centering
    \includegraphics[width=0.52\textwidth]{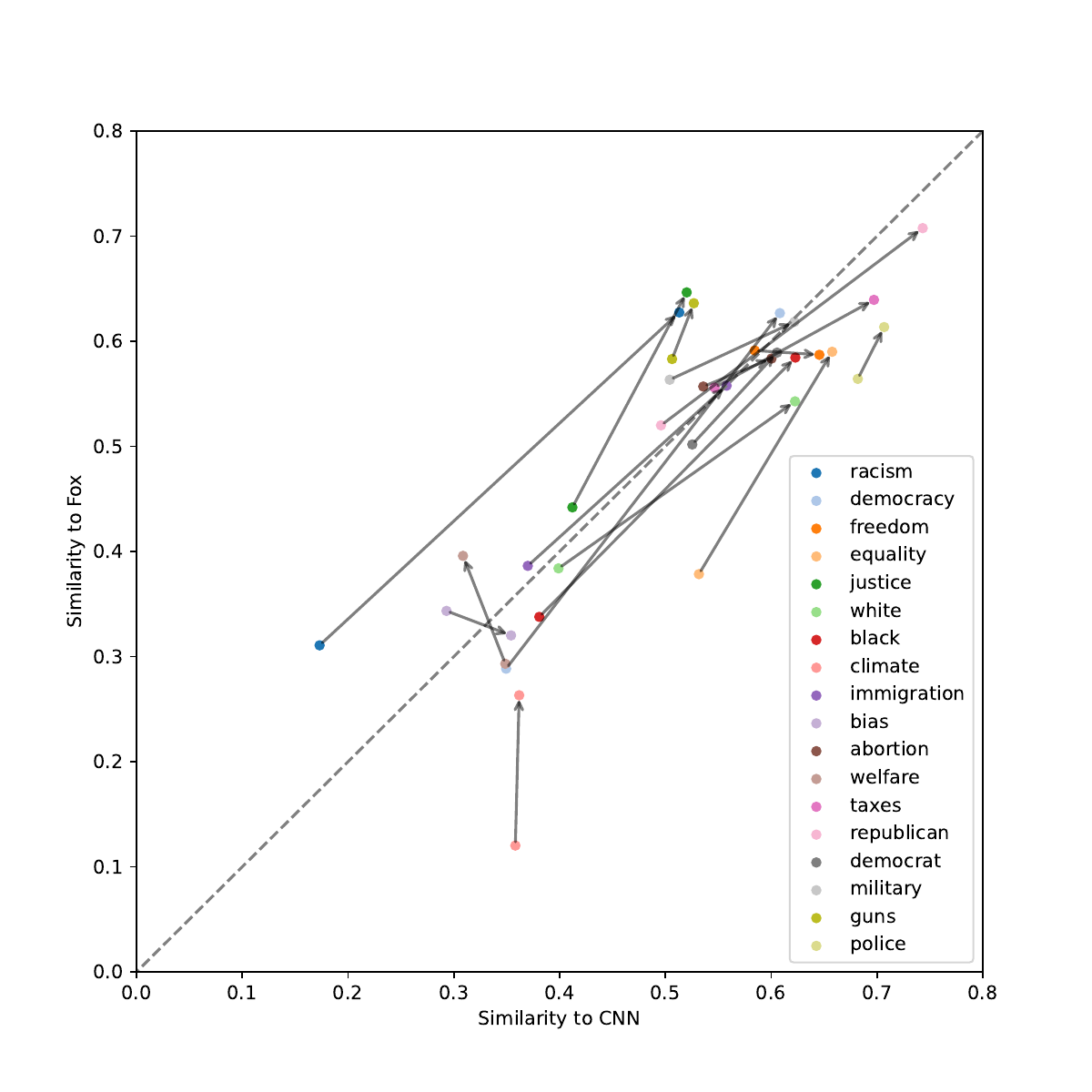}
    \caption{Comparing embedding similarity for our target words to embeddings for CNN and Fox News between the years 2014-2016. Arrows show the shift for embeddings trained on data before acquisition to embeddings trained on data after acquisition. There is a clear trend towards increasing nationalization-- similarity tends to increase to both national news outlets.}
    \label{fig:sims}
\end{figure}

\paragraph{Embedding Similarity}
\autoref{fig:sims} shows how similarity with our seed words changes from before purchase to after purchase.  We find that there is a general trend towards increasing similarity between our polarized news outlets and local news stations after acquisition by Sinclair, with the majority of our seed words showing increased similarity to both outlets.
Almost all the arrows direct up and to the right along the central diagonal line. This shift indicates that the usage of these words becomes more similar to both Fox and CNN after Sinclair purchase. Only three words, (``freedom,'' ``welfare,'' and ``bias''), decrease noticeably in similarity to one of the national broadcasters, and each increases in similarity to the other broadcaster.  Thus, there is a clear shift towards increasingly national language as opposed to local language. Unlike the nearest neighbor results, we limit both the national and local station data to 2014 through 2016, minimizing the effect that different news events during different time periods has on our analysis. We do not observe a shift towards either national outlet in particular. While prior work has observed right-wing slant in Sinclair-owned stations \cite{tryon-sinclair}, we suspect word embeddings are not sufficient to capture this trend, as they have a limited context window size. Fox and CNN have also been measured as less polarized before 2020 \cite{ding2023sem-polar}, suggesting that comparisons against these outlets may not be sufficient for capturing slant. Nevertheless, this figure demonstrates that after purchase by Sinclair, stations tend to use these words in contexts similarly to polarized national news stations.


\section{Discussion}
\paragraph{Connection to Communications Theory}
Communications scholars have identified agenda setting and framing as tools for influencing public opinion, which can be used to characterize media bias \citep{entman2007framing}.
While agenda-setting broadly encompasses ways the media reports on some events at the exclusion of others (e.g., \textit{what} topics are covered), framing involves highlighting specific aspects of a topic or event in order to promote a particular interpretation (e.g., \textit{how} topics are covered) \citep{mccombs1972agenda,entman2007framing}, though these two mechanisms are not always distinct \citep{mccombs2001convergence}. When we consider our analysis through this lens, we note second-order agenda-setting strategies in our results. Increasing national focus, as well as focus on polarizing political issues, primarily occurs through agenda-setting, which is evident in topic-level changes (\autoref{fig:stm-all-main}-\autoref{fig:stm-2016-main}) and politicized word usage (\autoref{tab:n_neighbors}). Our results reveal some evidence of framing changes in vocabulary shifts within topics (\autoref{tab:content-topics-beforeafter}), but future work targeting framing specifically is needed to fully explore these trends. In contrast to prior work \citep{tryon-sinclair}, we do not find clear evidence of right-wing slant: our results do not consistently show Sinclair ownership is associated with more similarity to Fox than CNN. This may be explained by several factors, such as differences in what content news stations highlight on YouTube, limited polarization in CNN and Fox in the years we focus on~\citep{ding2023sem-polar}, or inability of our methods to capture nuanced coverage differences over broad changes in topics. 
Future work targeting framing could aid in disentangling these factors \citep{entman2007framing}.


\paragraph{Implications for viewers}

In addition to framing and agenda setting, a third tenet of media's distribution of power and influence over public opinion is \textit{priming}: the effects agenda-setting and framing have on the audience \citep{entman2007framing}.
While our study does not measure priming effects, previous work has connected the influence of Sinclair to material changes in partisan voting and to changed opinions of politicians \cite{miho2018,levendusky2022local}. The evidence that we uncover of shifts in digital content after purchase by Sinclair offers insight into the possible mechanisms leading to public opinion changes associated with Sinclair ownership. 
Our establishment of YouTube videos as a data source for analyzing content shifts also offers opportunities to studying priming. In future work, comments on YouTube videos could offer a way to directly examine viewers responses to specific content. This data could also be crossed with other social media sources, such as what links are shared on other platforms.



\section{Conclusion}
Across all text analysis methods, we find consistent evidence that acquisition by Sinclair is associated with increased coverage of national and political news, often at the expense of conventional local news topics such as cooking or local sports.
Combined with the increasing closure of local news outlets, our results offer a grim picture of the decline of community-focused news in the U.S. We further demonstrate the usefulness of YouTube data in measuring and understanding this trending, thus highlighting opportunities for follow-up research.


\section{Limitations}
While we target a causal question (the impacts of Sinclair purchase), our work is observational, which reduces our ability to draw causal conclusions.
There are potential unobserved confounders, including general shifts in coverage over time across all news outlets, possibly driven by industry-wide efforts to attract views and increase engagement.
We do take steps to correct for industry trends over time, including segmenting data by time in Table \ref{tab:fightingwords} and constructing a tightly controlled paired analysis (in appendix section \ref{sec:paired-analysis}).
However, these settings require restricting to smaller subsets of the data, which limits the analyses we can conduct.
Our dataset generally contains imbalances which could impact results, e.g., the distribution across stations and time is uneven.
Overall, while the consistency of content shifts coinciding with the timing of Sinclair purchases strongly suggests a causal relationship, we cannot definitively rule out the possibility of further confounders.
Finally, although we draw from previous work, our interpretations somewhat rely on subjective and personal judgments about what words are politicized. These choices have an impact on our conclusions.


\bibliography{custom.bib}

\clearpage

\clearpage

\appendix 

\section{Appendix}

\subsection{Paired Analysis}\label{sec:paired-analysis}

This paper has considered data from the same stations before and after purchase by Sinclair, allowing for control over confounding variables such as differing political content between stations. However, this also introduces time as a significant confounder, as stations' coverage of various news stories will obviously vary over time as current events unfold. We conduct an additional analysis in which we more tightly control for possible news variance over time, aiming to isolate the effects of purchase. We choose two stations in our dataset with similar nearby stations which were never Sinclair affiliated. We choose KECI/KCFW/KTVM (YouTube @NBCMontana) with KPAX-TV (YouTube @kpaxmissoula) as the non-Sinclair affiliated channel, both in western Montana (MT), and WCYB (YouTube @wcyb5) with WJHL-TV (YouTube @WJHLtv11), both in the middle of the Virginia-Tennessee (VA-TN) border. For the non-Sinclair affiliated stations, we scrape the same number of videos as in the respective Sinclair affiliated channel, scraping videos closest to the videos in the Sinclair affiliated channels. We also subsample the CNN and Fox data in the same way, selecting the subset in each closest to the videos in the Sinclair affiliated channels.

A drawback of analyzing these paired stations is that local news stations may copy each other's content. It is therefore possible that Sinclair's purchase of a local station also impacts content posted on other local stations, violating the condition of no interference. Regardless, this analysis provides additional insight: in preceding analyses, interference is less of a concern, but controls for time are less strict. Similar trends in both settings would lend support to the conclusion that Sinclair purchases impact coverage of local topics.
\paragraph{Methods} We train STMs, as in \textbf{RQ1}. We train two new STMs on the transcripts from the two local MT stations, two local VA-TN stations, and subsampled CNN and Fox data. We used time, and the before/after and Sinclair affiliated/non-affiliated subsets as covariates. The first STM is trained with all data, and the second with data from before 2020, as we aimed to study the Sinclair effect without the dominant pandemic topic.

\begin{figure}[ht]
    \centering
    \includegraphics[width=1\linewidth]{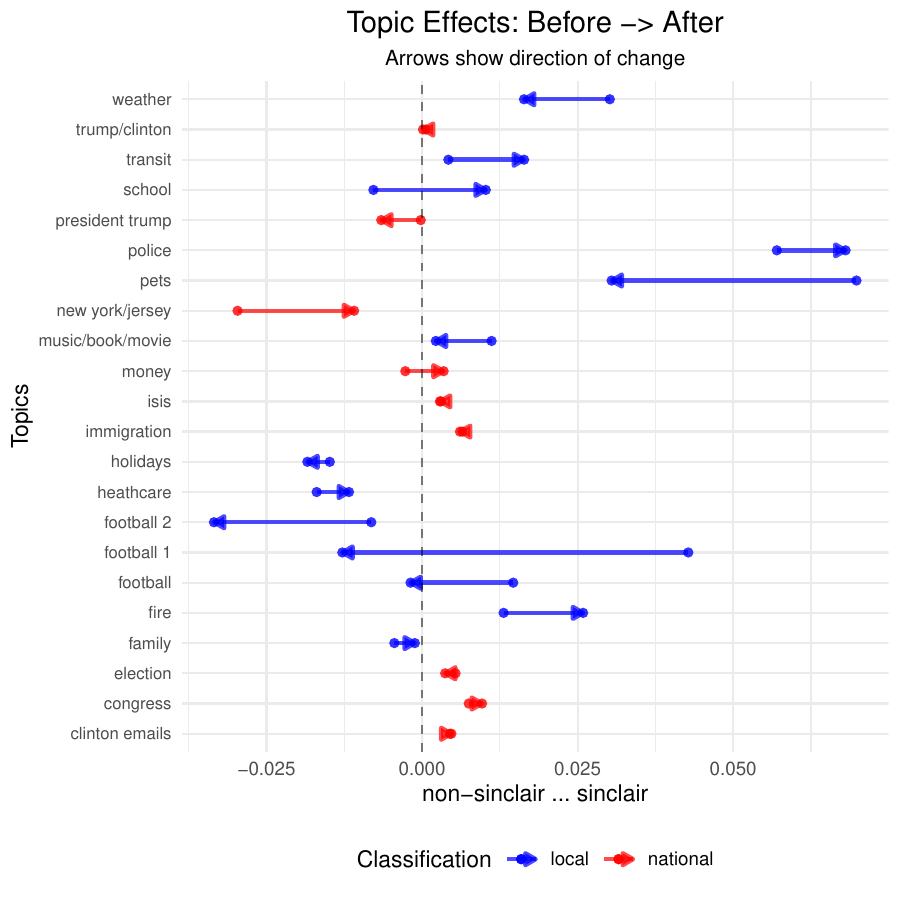}
    \caption{Results for STM on \textbf{paired data before 2020}. Change in topic proportion shifting from non-Sinclair to Sinclair affiliate on the x-axis, and shift before and after purchase date is shown with arrows. Red denotes national topics and blue denotes local topics. Topic list is shown on the y-axis. Topics with unclear national/local interpretation are omitted here, and included in Figure \ref{fig:paired-pre2020-appendix}.}
    \label{fig:paired-pre2020-main}
\end{figure}

\begin{figure*}
    \centering
    \includegraphics[width=0.9\linewidth]{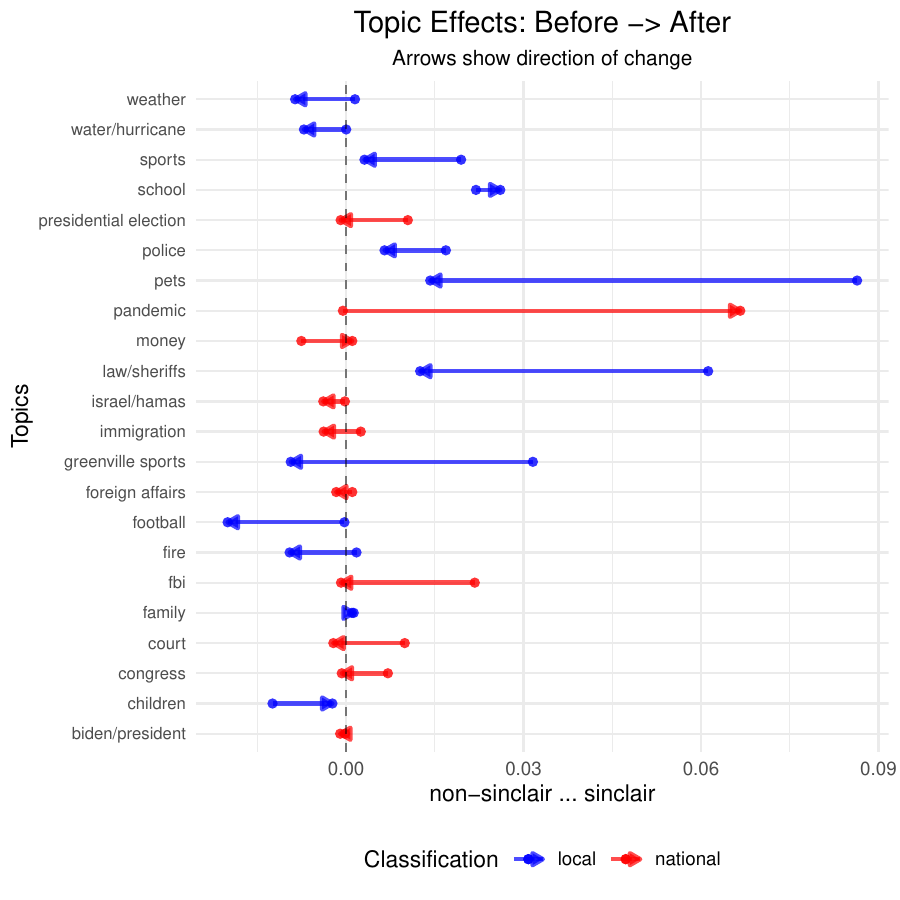}
    \caption{Results for STM on \textbf{all paired data}. Change in topic proportion shifting from non-Sinclair to Sinclair affiliate on the x-axis, and shift before and after purchase date is shown with arrows. Red denotes national topics and blue denotes local topics. Topic list is shown on the y-axis. Topics with unclear national/local interpretation are omitted here, and included in Figure \ref{fig:paired-all-appendix}.}
    \label{fig:paired-all-main}
\end{figure*}

\paragraph{Results}
STM results for selected topics are shown in Figures \ref{fig:paired-pre2020-main} and \ref{fig:paired-all-main}. As before, we manually assign representative names for each topic based on the full topics listed in Tables \ref{tab:topics-paired-all}-\ref{tab:topics-paired-pre2020}. We additionally assigned local/national/unclear topic labels. These plots display change in topic proportion between non-Sinclair affiliated channels and Sinclair affiliated channels before and after the purchase date of the Sinclair affiliated channel.

Topic modeling on all data demonstrates the prominence of pandemic coverage, which seems to dominate the topic proportion for dates after purchase, and is disproportionately covered by Sinclair owned channels. We also observe a shift away from local coverage for Sinclair-owned stations, which is likely due to Sinclair stations reporting less on local issues. STM results on paired data before 2020 demonstrate this same shift towards local coverage for non-Sinclair affiliated stations, especially seen in topics covering football and pets. We see less shift in national topics between channels, but do observe a small shift towards Sinclair-owned stations.

\subsection{Additional nearest neighbor results}
\begin{table*}[]
\begin{center}
\begin{tabular}{p{0.9in}|p{0.9in}||p{0.9in}|p{0.9in}||p{0.9in}|p{0.9in}}
 \rowcolor{gray!40} \multicolumn{2}{c}{\textbf{racism}} & \multicolumn{2}{c}{\textbf{democracy}}     & \multicolumn{2}{c}{\textbf{freedom}}  \\ \hline
\rowcolor{gray!25}Before       & After          & Before         & After          & Before         & After        \\ \hline\hline
racist     & racial          & equality     & extremism & freedoms  & democracy  \\
\rowcolor{gray!10}alleges    & injustice       & minister     & freedom   & sacrifice & freedoms   \\
appointed  & racist          & civil        & freedoms  & equality  & equality   \\
\rowcolor{gray!10}dismissed  & equality        & unrest       & nation    & pride     & sacred     \\
consulted  & rhetoric        & america      & values    & faithful  & liberation \\
\rowcolor{gray!10}pape       & africanamerican & conflict     & america   & nation    & slavery    \\
vulgar     & hatred          & islamic      & radical   & civil     & symbol     \\
\rowcolor{gray!10}derogatory & movement        & humanitarian & equality  & 1963      & birthright \\
goodell    & protesting      & muslim       & ideology  & democracy & nation     \\
\rowcolor{gray!10}judgment   & muslim          & nations      & defend    & rights    & ideals  \\ \hline
 \rowcolor{gray!40} \multicolumn{2}{c}{\textbf{justice}} & \multicolumn{2}{c}{\textbf{immigration}}     & \multicolumn{2}{c}{\textbf{welfare}}  \\ \hline
\rowcolor{gray!25}Before       & After          & Before         & After          & Before         & After        \\ \hline\hline
prosecution & judicial       & immigrants    & undocumented & claims   & services       \\
\rowcolor{gray!10}argued      & injustice      & reform        & immigrants   & petition & leno           \\
innocent    & criminal       & undocumented  & reform       & misuse   & westmoreland   \\
\rowcolor{gray!10}civil       & prosecute      & congress      & deportation  & claiming & mandated       \\
prosecutor  & innocence      & citizenship   & obamacare    & deny     & systematically \\
\rowcolor{gray!10}selfdefense & accountable    & nra           & repeal       & abuse    & lenor          \\
actions     & prejudice      & conservatives & deport       & seek     & medicaid       \\
\rowcolor{gray!10}conviction  & collective     & unaccompanied & repealing    & fraud    & stamps         \\
testify     & dignity        & border        & enact        & status   & cheri          \\
\rowcolor{gray!10}judicial    & accountability & bipartisan    & latinos      & coverup  & care   \\ \hline
 \rowcolor{gray!40} \multicolumn{2}{c}{\textbf{taxes}} & \multicolumn{2}{c}{\textbf{republican}}     & \multicolumn{2}{c}{\textbf{democrat}}  \\ \hline
\rowcolor{gray!25}Before       & After          & Before         & After          & Before         & After        \\ \hline\hline
tax       & tax          & democrat       & democratic   & republican  & democratic  \\
\rowcolor{gray!10}income    & income       & democratic     & democrats    & candidate   & delegate    \\
taxpayers & debt         & candidate      & gop          & democratic  & republican  \\
\rowcolor{gray!10}budgets   & corporations & grothman       & republicans  & incumbent   & incumbent   \\
debt      & costs        & representative & democrat     & reelection  & partisan    \\
\rowcolor{gray!10}rates     & revenue      & senator        & candidate    & campaigning & reelection  \\
pension   & wealthy      & congressman    & conservative & primary     & congressman \\
\rowcolor{gray!10}paying    & taxpayers    & reelection     & caucus       & nomination  & hollen      \\
fees      & burden       & politics       & partisan     & romney      & democrats   \\
\rowcolor{gray!10}fiscal    & deferral     & democrats      & electorate   & mitt        & candidate      \\ \hline
 \rowcolor{gray!40} \multicolumn{2}{c}{\textbf{military}} & \multicolumn{2}{c}{\textbf{guns}}     & \multicolumn{2}{c}{\textbf{police}}  \\ \hline
\rowcolor{gray!25}Before       & After          & Before         & After          & Before         & After        \\ \hline\hline
troops      & army        & weapons       & firearms      & authorities   & mpd         \\
\rowcolor{gray!10}iraq        & soldiers    & gun           & gun           & investigators & officers    \\
afghanistan & afghanistan & firearms      & handgun       & witnesses     & authorities \\
\rowcolor{gray!10}civilian    & combat      & rifles        & firearm       & detectives    & detectives  \\
combat      & troops      & ammunition    & semiautomatic & deputies      & juveniles   \\
\rowcolor{gray!10}navy        & marines     & firearm       & illegal       & mpd           & cops        \\
forces      & armys       & handguns      & weapons       & sources       & suspects    \\
\rowcolor{gray!10}soldier     & soldier     & semiautomatic & rifles        & officers      & patrols     \\
iraqi       & forces      & concealed     & handguns      & suspects      & deputies    \\
\rowcolor{gray!10}soldiers    & overseas    & rifle         & criminals     & suspect       & suspect \\
\end{tabular}
\caption{Nearest neighbors of other words. These words demonstrated no significant interpretable changes.}\label{tab:more-neighbors}
\end{center}
\end{table*}
\autoref{tab:more-neighbors} shows results for our remaining query words. We find that these results demonstrate less interpretable shifts in ideology or framing.

\subsection{Dataset example}

\begin{table*}
    \begin{center}
    \begin{tabular}{cp{0.8in}p{1in}p{0.8in}p{2.4in}}
         Channel & Title & URL & Date & Transcript  \\
         \hline
         wsbttv & Cold temperatures could impact fruit production & https://www.you tube.com/watch? v=6zEWpCJ95hc & 2024-03-24T16:00:11Z & typically Mother's Day is when apple orchards across the area start to see their crops in full bloom at ker Sunrise Orchards though they're already a month ahead of schedule with... \\
    \end{tabular}
    \caption{Example from the scraped YouTube data}
    \label{tab:example}
    \end{center}
\end{table*}
\autoref{tab:example} shows an example of how our data is structured.

\subsection{Topic Modeling Details}\label{appendix:topic-modeling}
Tables \ref{tab:topics-all}, \ref{tab:topics2014}, \ref{tab:topics2015}, and \ref{tab:topics2016} contain columns with words from various methods for evaluating top words for topics from a structured topic model.
Highest Prob denotes the highest probability words. 
FREX denotes the highest ranking FREX (FRequency and EXclusivity) words. 
Lift denotes the highest scoring words by lift, which weights words and divides by their frequency in other topics. This gives higher weight to words less frequent among other topics. 
Score denotes the best words by score by dividing the log frequency
of the word in the topic by the log frequency of the word in other topics.
The ``Label'' column contains our manually written labels for each topic.

\onecolumn

\renewcommand{\arraystretch}{1.2}
\rowcolors{2}{gray!10}{white}
\begin{center}


\begin{longtable}{>{\bfseries}c | >{\bfseries}m{0.08\textwidth} | m{0.15\textwidth} | m{0.15\textwidth} | m{0.15\textwidth} | m{0.15\textwidth}}

\caption{Topics for All-Time STM. See section appendix for column information.}\label{tab:topics-all} \\

\hline
\textbf{Topic} & \textbf{Label} & \textbf{Highest Prob} & \textbf{FREX} & \textbf{Lift} & \textbf{Score} \\
\hline
\endfirsthead

\hline
\textbf{Topic} & \textbf{Label} & \textbf{Highest Prob} & \textbf{FREX} & \textbf{Lift} & \textbf{Score} \\
\hline
\endhead

1 & police & police, fire, car, say, officers, officer, scene & fire, officer, scene, car, police, crash, officers & submit, firefighters, gunman, crash, transported, driver, flames & police, submit, officers, officer, car, scene, fire \\
2 & law/ crime & law, people, gun, crime, enforcement, violence, police & violence, guns, enforcement, illegal, gun, law, immigrants & climb, criminals, guns, firearms, marijuana, sanctuary, violence & climb, law, enforcement, crime, police, immigration, violence \\
3 & football & first, get, back, right, going, one, got & ball, yards, tennessee, quarter, touchdown, videos, play & videos, touchdown, yards, snap, tennessee, ball, bristol & videos, touchdown, ball, game, yards, coach, quarterback \\
4 & video & see, right, just, video, can, back, one & video, phone, saw, sir, correct, pictures, yes & video, footage, phone, recording, phones, images, cameras & video, phone, sir, yes, okay, correct, see \\
5 & city/ mayor & new, city, will, mayor, york, building, nbc & mayor, city, bridge, nbc, project, providence, council & champion, mayor, bridge, providence, construction, citys, city & champion, city, mayor, providence, montana, nbc, york \\
6 & court & court, case, judge, trump, attorney, justice, will & jury, judge, attorney, trial, hunter, court, indictment & testifying, indictment, jury, mar-alago, prosecution, lawyers, merrick & testifying, trump, court, hunter, jury, attorney, documents \\
7 & president & president, going, house, white, said, foreign, just & foreign, president, presidents, white, administration, congress, secretary & foreign, presidents, cabinet, president-elect, broadly, bipartisan, summit & foreign, president, congress, obama, presidents, administration, secretary \\
8 & fbi & information, department, government, fbi, security, committee, report & fbi, information, committee, chairman, agencies, agency, data & trusted, inspector, cyber, agency, breach, agencies, privacy & trusted, fbi, investigation, government, information, committee, intelligence \\
9 & - & gtgt, reporter, said, people, gtgtgt, dont, say & gtgt, reporter, gtgtgt, cnn, jake, plane, ten & gtgt, gtgtgt, liar, reporter, brooke, jake, cnn & gtgt, liar, gtgtgt, reporter, cnn, e-mail, e-mails \\
10 & cooking & just, little, going, can, okay, like, right & cheese, chicken, okay, cream, recipe, sauce, little & garlic, oven, recipes, sauce, chocolate, flavors, toss & toss, cheese, recipes, sauce, okay, garlic, gonna \\
11 & israel/ hamas & israel, ukraine, russia, war, military, iran, will & hamas, gaza, israel, israeli, putin, iran, russia & casualties, gaza, hamas, israelis, missiles, israeli, palestinians & ukraine, hamas, israel, casualties, russia, gaza, iran \\
12 & country & country, people, america, will, american, states, united & america, freedom, nation, rights, veterans, country, abortion & thus, values, liberty, freedom, religious, dignity, marriage & thus, america, country, americans, democracy, american, abortion \\
13 & clinton & clinton, hillary, shes, campaign, debate, said, obama & clinton, hillary, clintons, sanders, bernie, emails, campaign & wore, clintons, clinton, hillary, bernie, sanders, server & clinton, hillary, wore, clintons, sanders, obama, bernie \\
14 & joy & like, show, love, guy, greg, one, shes & greg, jesse, movie, laughter, guy, funny, film & joy, movie, movies, jeanine, jesse, greg, song & joy, greg, jesse, laughter, movie, love, jeanine \\
15 & - & know, think, going, dont, thats, like, just & know, think, mean, dont, youre, thats, theres & gosh, mean, neil, nobodys, know, think, youre & gosh, know, think, mean, going, dont, youre \\
16 & biden & biden, joe, border, house, democrats, president, republicans & biden, joe, border, mccarthy, bidens, democrats, harris & maga, ainsley, pelosi, kamala, kayleigh, newsom, mccarthy & biden, maga, border, democrats, republicans, bidens, joe \\
17 & money & money, million, dollars, pay, tax, bill, state & dollars, tax, pay, money, budget, taxes, million & friendship, taxpayers, income, revenue, tax, dollars, budget & friendship, tax, money, dollars, budget, taxes, million \\
18 & isis & isis, attack, iraq, syria, now, attacks, war & isis, iraq, syria, terror, terrorist, terrorism, muslim & islam, isis, brutal, qaeda, syrian, refugees, iraqi & isis, brutal, syria, iraq, terrorist, terror, islamic \\
19 & station & news, now, says, county, live, tonight, today & abc, maryland, metro, wsbt, county, bend, news & heal, suzanne, fairfax, allison, arlington, patrice, georges & county, heal, wsbt, abc, news, metro, fairfax \\
20 & health & health, can, care, get, medical, now, hospital & cancer, doctor, doctors, disease, patients, health, medical & disease, ear, diagnosed, symptoms, virus, doctors, vaccine & ear, health, patients, cancer, disease, hospital, vaccine \\
21 & trump & trump, donald, hes, republican, think, election, going & voters, candidates, polls, trump, donald, iowa, republican & odd, rubio, marco, electorate, rnc, mitt, romney & trump, donald, odd, republican, voters, election, republicans \\
22 & china & china, now, energy, world, company, new, chinese & energy, china, market, climate, prices, chinese, industry & aim, pipeline, industry, prices, consumers, electric, markets & aim, china, chinese, prices, economy, inflation, climate \\
23 & weather & going, see, will, now, right, weather, well & snow, storm, weather, temperatures, rain, storms, winds & thunderstorms, temperatures, cloudy, crew, storms, showers, snow & crew, snow, temperatures, storms, rain, weather, storm \\
24 & station 2 & morning, year, fox, green, people, well, can & bay, christmas, green, holiday, wisconsin, rachel, event & tap, zoo, museum, birds, peterson, parade, fishing & tap, bay, appleton, oshkosh, fox, wisconsin, christmas \\
25 & school & school, kids, students, children, parents, schools, high & students, campus, school, schools, teachers, student, parents & alan, campus, campuses, teacher, superintendent, teachers, students & school, students, alan, kids, schools, parents, student \\
26 & sports & game, team, year, play, one, win, season & sports, game, games, fans, players, football, team & exclusive, notre, dame, tournament, nfl, baseball, irish & exclusive, game, notre, dame, football, coach, sports \\
27 & crime & found, say, case, man, two, -year-old, death & murder, -year-old, arrested, prison, victim, sexual, charged & cop, sexually, sentenced, dna, sexual, allegedly, murder & cop, murder, investigators, police, charges, arrested, -year-old \\
28 & comm-unity & can, need, work, people, will, community, make & community, help, work, rhode, working, resources, need & medicine, rhode, resources, partnership, community, nonprofit, resource & medicine, community, rhode, thank, need, communities, families \\
29 & media & media, people, news, fox, said, saying, story & media, twitter, racist, social, post, youtube, tucker & usa, musk, racist, elon, carlson, twitter, tucker & usa, media, twitter, racist, musk, fox, elon \\
30 & family/ church & family, life, just, years, day, know, time & family, life, father, church, loved, mom, friends & hats, pastor, sisters, grandfather, pray, remembered, jesus & hats, family, father, life, son, mom, mother \\
\hline
\end{longtable}
\end{center}

\renewcommand{\arraystretch}{1.2}
\rowcolors{2}{gray!10}{white}
\begin{center}


\begin{longtable}{>{\bfseries}c | >{\bfseries}m{0.08\textwidth} | m{0.15\textwidth} | m{0.15\textwidth} | m{0.15\textwidth} | m{0.15\textwidth}}

\caption{Topics for 2014 STM. See appendix for column information.}\label{tab:topics2014} \\

\hline
\textbf{Topic} & \textbf{Label} & \textbf{Highest Prob} & \textbf{FREX} & \textbf{Lift} & \textbf{Score} \\
\hline
\endfirsthead

\hline
\textbf{Topic} & \textbf{Label} & \textbf{Highest Prob} & \textbf{FREX} & \textbf{Lift} & \textbf{Score} \\
\hline
\endhead

1 & isis & isis, iraq, military, syria, united, israel, will & isis, syria, israel, iraq, hamas, gaza, israeli & hamas, iraqi, islamic, joy, qaeda, syria, syrian & isis, joy, iraq, syria, hamas, gaza, israel \\
2 & movie & north, movie, film, korea, show, kim, theater & korea, movie, movies, film, theater, kim, hollywood & brutal, theaters, korea, movies, cyber, comedy, movie & brutal, movie, korea, film, hollywood, north, theaters \\
3 & ukraine/ russia & ukraine, russia, russian, putin, bridge, key, president & ukraine, russia, russian, putin, calm, bridge, ukrainian & calm, putin, russia, russian, russians, ukraine, ukrainian & calm, ukraine, russia, russian, putin, ukrainian, sanctions \\
4 & video & video, phone, see, security, camera, call, cell & video, phone, cell, camera, cameras, tape, ray & video, cell, ray, roger, phones, phone, cameras & video, phone, nfl, cell, cameras, camera, surveillance \\
5 & court & court, case, says, said, today, attorney, now & judge, court, attorney, documents, prosecutors, trial, guilty & los, courtroom, judge, allegations, pleaded, sentenced, lawsuit & los, court, attorney, judge, prosecutors, charges, investigation \\
6 & nbc & last, two, one, three, night, ago, years & nbc, last, night, island, three, ago, providence & unlikely, patrice, nbc, providence, susie, tony, frank & unlikely, nbc, providence, rhode, island, last, night \\
7 & health/ ebola & health, ebola, care, hospital, medical, now, patients & ebola, patients, disease, health, doctors, patient, virus & patients, professionals, cdc, ebola, outbreak, virus, infected & ebola, professionals, health, hospital, patients, virus, disease \\
8 & city/ mayor & city, will, new, says, mayor, building, now & mayor, city, council, project, property, marijuana, construction & authority, mayor, mayors, council, marijuana, city, construction & authority, city, mayor, council, project, property, marijuana \\
9 & shopping & can, like, one, use, get, just, new & shop, buy, store, stores, sell, items, products & spicy, app, products, stores, shop, shopping, plastic & spicy, store, can, products, shop, stores, food \\
10 & police & police, say, man, now, county, live, tonight & suspect, prince, police, victim, investigators, -year-old, georges & basement, stabbed, detectives, plater, roz, year-old, gunman & police, basement, county, investigators, suspect, victim, abc \\
11 & - & think, people, dont, thats, going, want, can & think, question, welcome, dont, understand, sort, talk & welcome, frankly, perspective, shouldnt, agree, necessarily, legitimate & welcome, think, people, dont, question, want, youre \\
12 & money & money, state, million, dollars, pay, company, business & jobs, million, tax, money, dollars, pay, budget & chinese, manufacturing, taxpayers, taxes, taxpayer, tax, jobs & chinese, dollars, tax, money, million, jobs, taxes \\
13 & weather & water, ice, says, winter, river, power, lake & ice, warning, water, river, trees, fish, boat & warning, dnr, boats, fishing, swimming, flooding, trees & warning, water, ice, winter, lake, fish, trees \\
14 & cooking & going, just, little, okay, can, like, really & recipe, sauce, cheese, flavor, chicken, cream, butter & teaspoon, butter, flavor, flour, onion, onions, oven & mustard, sauce, recipe, cheese, recipes, flavor, cream \\
15 & - & know, like, just, really, think, got, dont & know, mean, yeah, like, ive, really, hes & wow, weird, nervous, mean, know, guess, magic & wow, know, yeah, like, think, hes, mean \\
16 & family & family, life, just, children, son, shes, child & son, family, father, child, mother, life, daughter & sleep, father, son, brother, sister, child, daughter & sleep, family, child, son, mother, children, father \\
17 & wsbt & county, says, south, wsbt, channel, bend, kelly & wsbt, bend, joseph, dog, elkhart, desk, channel & wsbts, registered, fillmore, wsbs, denise, elkhart, joseph & registered, county, wsbt, bend, elkhart, kelly, joseph \\
18 & football & game, team, play, one, season, win, football & football, game, sports, players, games, win, team & congratulations, playoffs, touchdown, redskins, quarterback, playoff, championship & congratulations, game, football, notre, dame, players, coach \\
19 & events & fox, green, people, bay, year, event, wisconsin & tickets, annual, fox, event, packers, museum, music & champion, ronaldo, donation, tickets, organizers, annual, parade & champion, fox, bay, green, packers, wisconsin, appleton \\
20 & media & new, story, women, media, york, news, show & york, media, book, wrote, twitter, women, read & dice, diana, tweeted, magazine, twitter, tweet, york & dice, media, women, york, book, gtgtgt, twitter \\
21 & weather 2 & snow, weather, tomorrow, will, morning, going, now & snow, wind, rain, degrees, temperatures, weather, winds & chills, cloudy, snow, wind, showers, forecast, meteorologist & chills, snow, temperatures, rain, weather, degrees, storm \\
22 & morning & well, good, right, morning, yeah, can, just & yeah, fun, pauline, cool, good, morning, yes & chili, pauline, zoo, garden, emily, awesome, deem & chili, pauline, fun, yeah, morning, cool, awesome \\
23 & school & school, students, high, kids, schools, college, program & students, school, schools, student, college, campus, teachers & students, classroom, materials, teachers, elementary, teacher, superintendent & school, students, materials, schools, student, kids, teachers \\
24 & police 2 & police, officer, officers, gun, brown, shot, michael & officer, gun, ferguson, officers, enforcement, wilson, michael & convenience, ferguson, officer, guns, cop, missouri, louis & police, convenience, officer, officers, ferguson, jury, gun \\
25 & veterans & world, country, will, today, american, years, people & veterans, america, nation, honor, world, american, church & americas, veterans, nation, veteran, vietnam, honor, sergeant & americas, veterans, war, world, american, america, afghanistan \\
26 & reporter & gtgt, reporter, said, say, dont, people, yes & gtgt, reporter, gtgtgt, cnn, yes, listen, didnt & reporter, gtgt, cnns, gtgtgt, cnn, brooke, anderson & reporter, gtgt, gtgtgt, cnn, cnns, happened, said \\
27 & president & president, house, republicans, obama, republican, will, democrats & republicans, republican, democrats, clinton, senate, hillary, election & agenda, republicans, democrats, republican, democrat, hillary, immigration & agenda, republicans, president, democrats, republican, hillary, clinton \\
28 & fire & fire, car, road, morning, just, live, now & fire, cars, driver, road, car, truck, drivers & flames, firefighters, driver, intersection, lanes, brianne, firefighter & flames, fire, car, firefighters, driver, cars, crash \\
29 & plane & plane, flight, air, search, airport, information, area & plane, flight, aircraft, ocean, airport, pilot, ship & malaysian, aviation, islands, malaysia, plane, flight, wreckage & plane, islands, flight, aircraft, malaysian, search, airlines \\
30 & states & will, going, now, right, line, see, get & line, california, zone, virginia, train, space, station & zone, trains, california, ring, mexico, train, line & zone, virginia, line, space, going, california, station \\
\hline
\end{longtable}

\end{center}

\renewcommand{\arraystretch}{1.2}
\rowcolors{2}{gray!10}{white}
\begin{center}


\begin{longtable}{>{\bfseries}c | >{\bfseries}m{0.08\textwidth} | m{0.15\textwidth} | m{0.15\textwidth} | m{0.15\textwidth} | m{0.15\textwidth}}

\caption{Topics for 2015 STM. See appendix for column information.}\label{tab:topics2015} \\

\hline
\textbf{Topic} & \textbf{Label} & \textbf{Highest Prob} & \textbf{FREX} & \textbf{Lift} & \textbf{Score} \\
\hline
\endfirsthead

\hline
\textbf{Topic} & \textbf{Label} & \textbf{Highest Prob} & \textbf{FREX} & \textbf{Lift} & \textbf{Score} \\
\hline
\endhead

1 & planned parenthood & phone, planned, officer, body, parenthood, cell, fired & planned, phone, parenthood, cell, videos, fired, camera & submit, planned, parenthood, footage, images, phone, videos & parenthood, officer, submit, planned, phone, cell, videos \\
2 & video & video, car, shot, stop, saw, man, pulled & video, van, pulled, car, shot, bus, leg & video, van, recording, yelling, screaming, belt, leg & video, car, van, shot, pulled, -year-old, neck \\
3 & police & police, officers, black, gun, city, officer, community & gun, baltimore, black, officers, gray, guns, violence & ferguson, policing, recover, freddie, protests, baltimore, gray & police, officers, recover, officer, baltimore, gun, black \\
4 & congress & house, white, congress, bill, will, republicans, senate & congress, senate, speaker, house, vote, bill, legislation & speaker, con, boehner, bipartisan, lawmakers, veto, shutdown & senate, con, congress, republicans, democrats, republican, vote \\
5 & cancer & women, children, can, health, kids, child, cancer & cancer, disease, doctor, brain, study, doctors, health & cancer, disease, grace, symptoms, medication, diagnosed, brain & grace, cancer, health, disease, patients, children, child \\
6 & animals & year, home, people, come, day, dog, event & dog, dogs, christmas, owner, animal, store, animals & furniture, pets, pet, donate, dogs, animals, animal & furniture, dog, dogs, store, animal, dollars, year \\
7 & obama & president, obama, policy, foreign, speech, administration, world & foreign, policy, obama, president, obamas, presidents, speech & web, foreign, obamas, policy, obama, president, oval & president, obama, web, foreign, policy, barack, speech \\
8 & station & now, live, news, city, will, new, abc & metro, maryland, rhode, island, sam, providence, abc & sale, bowser, rhode, trains, providence, sweeney, buses & sale, metro, providence, city, rhode, county, abc \\
9 & fun & fun, anybody, join, dance, joy, laugh, cheering & fun, join, joy, anybody, dance, laugh, cheering & joy, join, fun, cheering, laugh, dance, guide & fun, joy, join, anybody, dance, laugh, cheering \\
10 & trump & trump, donald, hes, republican, carson, bush, new & trump, donald, carson, polls, iowa, stage, poll & stage, carsons, trump, donald, trumps, romney, carson & stage, trump, donald, carson, jeb, republican, bush \\
11 & - & gtgt, reporter, said, gtgtgt, get, dont, like & gtgt, reporter, gtgtgt, cnn, yes, ten, jake & reporter, gtgt, gtgtgt, translator, cnns, jake, don & reporter, gtgt, gtgtgt, cnn, e-mail, translator, jake \\
12 & weather & will, water, morning, snow, tomorrow, see, day & snow, weather, water, rain, storm, temperatures, degrees & futurecast, rain, inches, showers, snow, thunderstorms, clouds & snow, temperatures, rain, weather, showers, water, degrees \\
13 & tech-nology & new, government, information, federal, use, company, can & technology, data, company, employees, cyber, systems, companies & lowest, consumers, technology, cyber, users, systems, data & lowest, government, data, cyber, federal, technology, company \\
14 & football & game, team, one, play, year, season, football & game, football, sports, games, notre, dame, team & cubs, nfl, playoff, tournament, coaches, irish, playoffs & cubs, notre, game, dame, sports, football, irish \\
15 & money & money, million, tax, dollars, pay, cut, jobs & tax, cut, money, taxes, million, jobs, pay & cut, tax, taxes, growth, taxpayers, revenue, income & cut, tax, money, taxes, dollars, economy, budget \\
16 & - & people, can, will, think, want, need, make & important, things, process, need, sure, talk, forward & inner, newstalk, challenges, bruce, decisions, discussions, conversations & inner, people, think, need, important, will, can \\
17 & cooking & can, just, okay, right, like, yeah, little & okay, yeah, gonna, cheese, nice, cool, eat & cheese, oven, recipes, salad, butter, chocolate, cooking & recipes, yeah, okay, gonna, cheese, chocolate, great \\
18 & - & know, going, right, well, thats, get, just & know, theyre, mean, going, really, lot, yeah & wave, know, sort, theyre, mean, whats, probably & know, wave, mean, going, right, theyre, think \\
19 & family & like, just, years, family, life, one, love & movie, friends, amazing, book, loved, love, dad & justin, song, diana, movie, instagram, movies, sing & justin, movie, film, love, book, family, music \\
20 & school & school, wsbt, students, south, says, county, bend & students, wsbt, bend, school, schools, indiana, joseph & students, copeland, com, crenshaw, elkart, fillmore, teachers & wsbt, com, school, students, bend, elkhart, county \\
21 & debate & debate, candidates, think, night, last, rubio, governor & debate, walker, carly, candidates, debates, rubio, marco & cnbc, moderators, winners, ferina, debates, karly, carly & debate, rubio, candidates, winners, carly, marco, debates \\
22 & presid-ential candidates & clinton, hillary, shes, sanders, campaign, democratic, state & sanders, clinton, biden, hillary, bernie, clintons, server & biden, sanders, server, clintons, hearings, bernie, emails & hillary, clinton, biden, sanders, clintons, bernie, hearings \\
23 & immig-ration & people, country, governor, law, states, immigration, going & immigration, illegal, governor, border, law, country, laws & citizenship, creation, undocumented, latino, mexico, amnesty, illegal & immigration, governor, creation, law, border, immigrants, citizenship \\
24 & police & police, fire, now, say, just, one, scene & scene, fire, driver, injuries, hospital, injured, firefighters & flames, firefighters, gunshots, suv, scene, transported, driver & police, flames, scene, hospital, county, investigators, injuries \\
25 & isis & isis, military, iran, will, iraq, deal, syria & nuclear, iran, troops, russia, assad, military, forces & sanctions, iranians, irans, assad, kurds, ukraine, troops & irans, isis, iran, syria, iraq, nuclear, assad \\
26 & plane & officials, information, one, security, plane, now, may & plane, flight, pilot, airport, officials, search, sources & bodies, airlines, flight, pilots, plane, pilot, drone & bodies, flight, plane, investigation, passengers, aircraft, airport \\
27 & court & case, court, judge, attorney, said, today, charges & charges, judge, prison, jury, attorney, charged, trial & hernandez, courtroom, sentenced, jurors, prosecutors, jury, sentencing & hernandez, court, charges, attorney, jury, murder, prosecutors \\
28 & media & think, dont, said, say, know, people, hes & media, dont, guy, mean, think, doesnt, agree & curious, journalists, media, stupid, offended, apologize, ridiculous & curious, think, media, dont, know, mean, hes \\
29 & terrorism & isis, attack, people, paris, attacks, terror, terrorism & paris, refugees, terrorists, islam, terror, terrorist, muslims & massacre, jihad, paris, radicalized, refugees, islam, bernardino & isis, refugees, muslims, paris, muslim, massacre, islam \\
30 & church & people, church, pope, faith, religious, today, rights & pope, marriage, faith, church, flag, gay, religious & introduced, pope, -sex, gay, marriage, cuba, bible & introduced, pope, church, religious, marriage, faith, gay\\ 
\hline

\end{longtable}
\end{center}

\renewcommand{\arraystretch}{1.2}
\rowcolors{2}{gray!10}{white}
\begin{center}


\begin{longtable}{>{\bfseries}c | >{\bfseries}m{0.08\textwidth} | m{0.15\textwidth} | m{0.15\textwidth} | m{0.15\textwidth} | m{0.15\textwidth}}

\caption{Topics for 2016 STM. See appendix for column information.}\label{tab:topics2016} \\

\hline
\textbf{Topic} & \textbf{Label} & \textbf{Highest Prob} & \textbf{FREX} & \textbf{Lift} & \textbf{Score} \\
\hline
\endfirsthead

\hline
\textbf{Topic} & \textbf{Label} & \textbf{Highest Prob} & \textbf{FREX} & \textbf{Lift} & \textbf{Score} \\
\hline
\endhead

1 & police & police, black, officers, gun, officer, shot, community & gun, officers, officer, police, charlotte, shooting, shot & submit, shootings, cops, gun, charlotte, officers, tulsa & police, officers, officer, gun, submit, charlotte, shooting \\
2 & govern-ment & government, security, will, federal, department, new, information & agencies, data, cyber, federal, agency, services, government & rent, recommendations, management, veto, agencies, lawmakers, software & rent, federal, government, cyber, security, agencies, department \\
3 & plane & air, plane, train, one, space, force, new & plane, unbelievable, train, flight, flying, space, air & unbelievable, pilot, airplane, jet, plane, landing, profile & unbelievable, plane, flight, train, aircraft, air, passengers \\
4 & family & just, like, family, life, one, kids, years & mother, son, father, kids, family, daughter, mom & music, elementary, mom, joy, mother, daughters, girl & music, kids, mother, family, parents, father, children \\
5 & - & way, different, make, line, get, trying, put & way, different, line, sometimes, ways, gets, rules & way, bottom, line, different, impression, useful, sometimes & way, different, rules, line, sometimes, ways, gets \\
6 & court & case, video, court, judge, will, evidence, justice & video, judge, attorney, case, charges, court, charged & video, jury, guilty, sentenced, lawsuit, trial, charged & video, attorney, judge, court, justice, charges, jury \\
7 & congress & president, house, party, republican, republicans, obama, democrats & senate, ryan, house, democrats, republicans, paul, party & pelosi, richard, mcconnell, senate, ryans, reid, schumer & richard, democrats, senate, republicans, republican, president, obama \\
8 & presid-ential candidates & trump, clinton, hillary, donald, debate, shes, think & debate, shes, debates, playing, hillary, candidates, clinton & moderator, moderators, playing, holt, debates, debate, lester & playing, clinton, trump, hillary, donald, debate, debates \\
9 & football & game, year, team, one, back, tonight, will & season, game, football, larry, sports, games, notre & championship, dame, limited, notre, larry, irish, sports & limited, notre, dame, football, game, players, sports \\
10 & - & gtgt, reporter, cnn, gtgtgt, campaign, one, says & reporter, present, gtgtgt, cnn, gtgt, cnns, tapper & present, reporter, tapper, cnns, sara, gtgtgt, aides & gtgt, reporter, present, gtgtgt, cnn, trumps, cnns \\
11 & weather & now, will, just, right, county, south, city & county, snow, storm, wsbt, bend, weather, road & hurricane, meteorologist, slow, snow, danielle, suzanne, inches & slow, wsbt, county, snow, elkhart, bend, storm \\
12 & isis & isis, war, syria, iraq, military, now, forces & isis, iraq, syria, troops, islamic, terrorists, syrian & baghdad, pentagon, caliphate, fighters, isil, iraqi, mosul & pentagon, isis, syria, iraq, iraqi, mosul, syrian \\
13 & family/ church & women, men, woman, born, said, bill, say & women, born, birth, sexual, church, christmas, christian & delivered, jewish, birth, marriage, abortion, certificate, pope & delivered, women, christmas, israel, born, church, abortion \\
14 & - & going, know, people, well, want, get, great & great, thank, going, know, want, ive, youve & appointment, fantastic, hopefully, appreciate, vets, luck, thank & appointment, going, know, people, great, thank, well \\
15 & foreign defense & russia, iran, north, nuclear, defense, putin, korea & defense, putin, korea, iran, russia, nuclear, sanctions & koreas, putin, sailors, defense, irans, jong-un, korea & defense, russia, korea, iran, nuclear, putin, russian \\
16 & trump & trump, donald, hes, trumps, campaign, president-elect, president & president-elect, pence, mike, romney, transition, mitt, trump & fortune, gingrich, bannon, swamp, pence, mattis, newt & trump, donald, president-elect, romney, fortune, trumps, pence \\
17 & - & gtgt, said, dont, people, think, say, want & gtgt, dont, jake, said, didnt, yes, listen & buddy, jake, gtgt, anderson, wolf, inappropriate, corey & gtgt, buddy, jake, think, gtgtgt, people, dont \\
18 & america & will, country, people, america, american, americans, make & applause, america, cheers, education, class, nation, inner & allen, applause, cheers, inequality, poverty, education, wage & applause, allen, cheers, america, jobs, hillary, country \\
19 & - & know, dont, think, like, right, thats, just & mean, yeah, dont, know, okay, youre, like & blacks, tucker, cuz, neil, weird, yeah, kimberly & blacks, think, mean, yeah, know, dont, hes \\
20 & us/race & people, country, speech, american, say, want, america & racist, flag, amendment, constitution, hate, muslim, freedom & web, anthem, racist, bigot, flag, liberty, amendment & web, racist, flag, speech, muslim, constitution, amendment \\
21 & media & media, news, new, york, press, fox, now & media, press, university, campus, coverage, fox, students & balanced, journalism, journalists, howard, campus, media, buzz & media, balanced, students, campus, fox, press, news \\
22 & - & think, really, lot, well, theres, thats, see & sort, really, think, kind, terms, certainly, obviously & luxury, sort, brexit, perspective, terms, reflection, broader & think, luxury, sort, really, lot, things, kind \\
23 & delegates & trump, sanders, cruz, donald, bernie, ted, republican & cruz, ted, sanders, bernie, delegates, rubio, convention & con, cruz, ted, delegates, kasich, caucuses, rubio & cruz, sanders, delegates, con, bernie, ted, trump \\
24 & terrorism & now, attack, new, two, people, one, police & authorities, brussels, injured, bomb, paris, suspect, attack & com, rahami, belgian, attacker, bomber, suspects, plot & com, police, brussels, fbi, investigators, terror, injured \\
25 & immig-ration & immigration, country, wall, illegal, going, border, will & immigration, illegal, border, immigrants, sanctuary, mexico, mexican & undocumented, deport, deportation, factory, sanctuary, aliens, immigration & immigration, factory, immigrants, illegal, border, sanctuary, mexico \\
26 & election & trump, vote, clinton, election, voters, states, hillary & voting, polls, electoral, votes, pennsylvania, ohio, poll & stops, electoral, stein, jill, ballots, battleground, recount & stops, trump, clinton, polls, vote, voters, electoral \\
27 & health & health, can, water, care, medical, now, get & doctor, medical, health, pneumonia, doctors, insurance, cancer & improvement, patients, symptoms, doctor, disease, diagnosed, doctors & health, improvement, pneumonia, patients, medical, doctor, doctors \\
28 & clinton emails & clinton, hillary, fbi, foundation, emails, information, state & emails, foundation, classified, server, fbi, email, e-mails & deliberately, server, classified, podesta, emails, wikileaks, comey & clinton, fbi, emails, classified, server, e-mails, hillary \\
29 & money & money, tax, million, jobs, business, going, pay & tax, money, trade, taxes, dollars, companies, market & gate, tax, prices, trillion, audit, market, rates & tax, gate, money, taxes, jobs, dollars, trade \\
30 & cuba & president, united, states, obama, world, will, american & united, cuba, prime, states, president, minister, castro & cuba, permanent, castro, prime, british, britain, communist & permanent, president, united, cuba, obama, castro, states \\
\hline
\end{longtable}
\end{center}

\newpage 
\begin{figure*}
    \centering
    \includegraphics[width=0.9\linewidth]{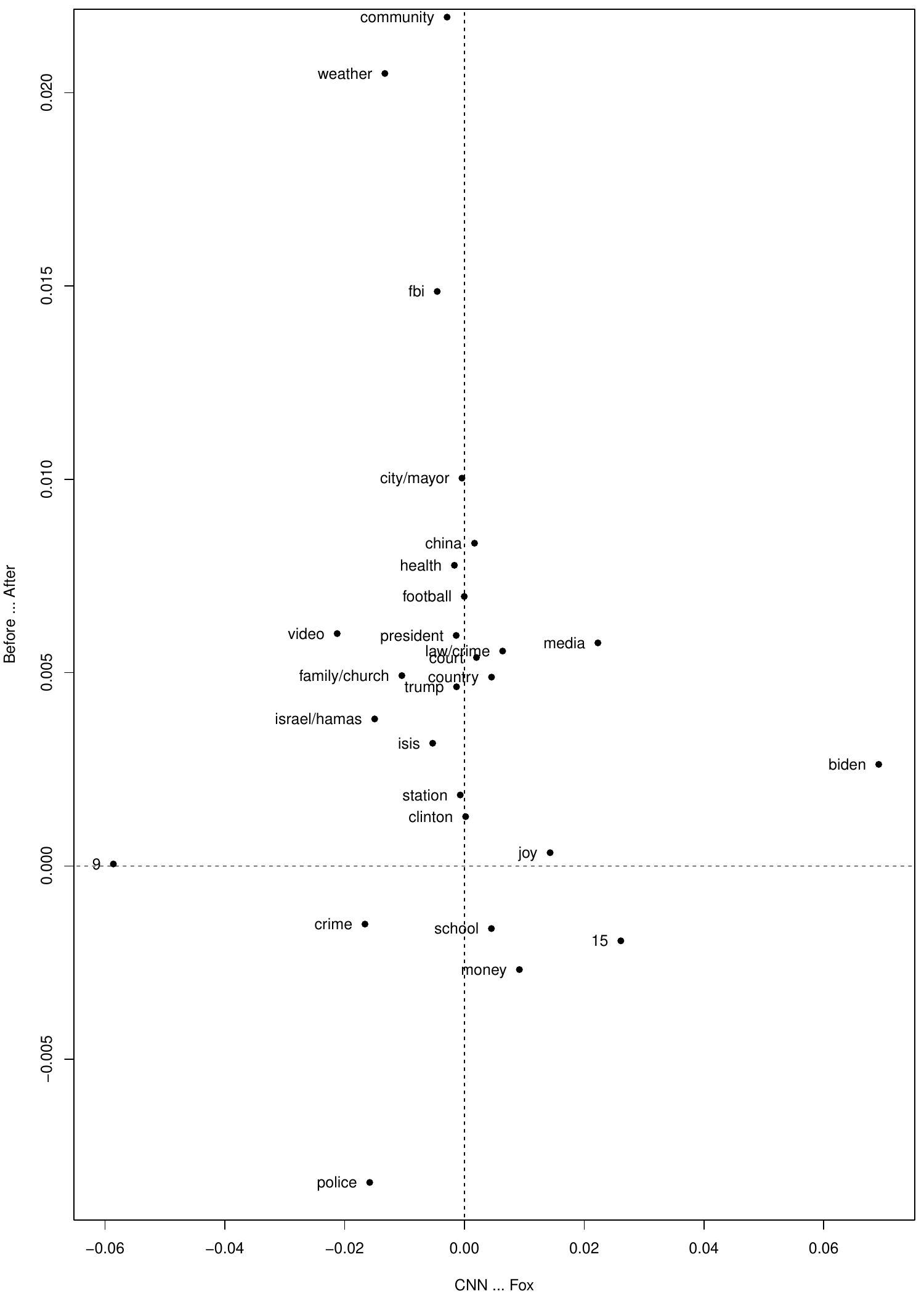}
    \caption{Results for STM on \textbf{all data}. Change in topic proportion shifting from CNN to Fox on the x-axis and before Sinclair purchase to after Sinclair purchase on the y-axis. Topics 10 (cooking), 26 (family), and 24 (station 2) are omitted, but can be found in Figure \ref{fig:stm-all-main}.}
    \label{fig:stm-all-appendix}
\end{figure*}

\newpage 
\begin{figure*}
    \centering
    \includegraphics[width=0.9\linewidth]{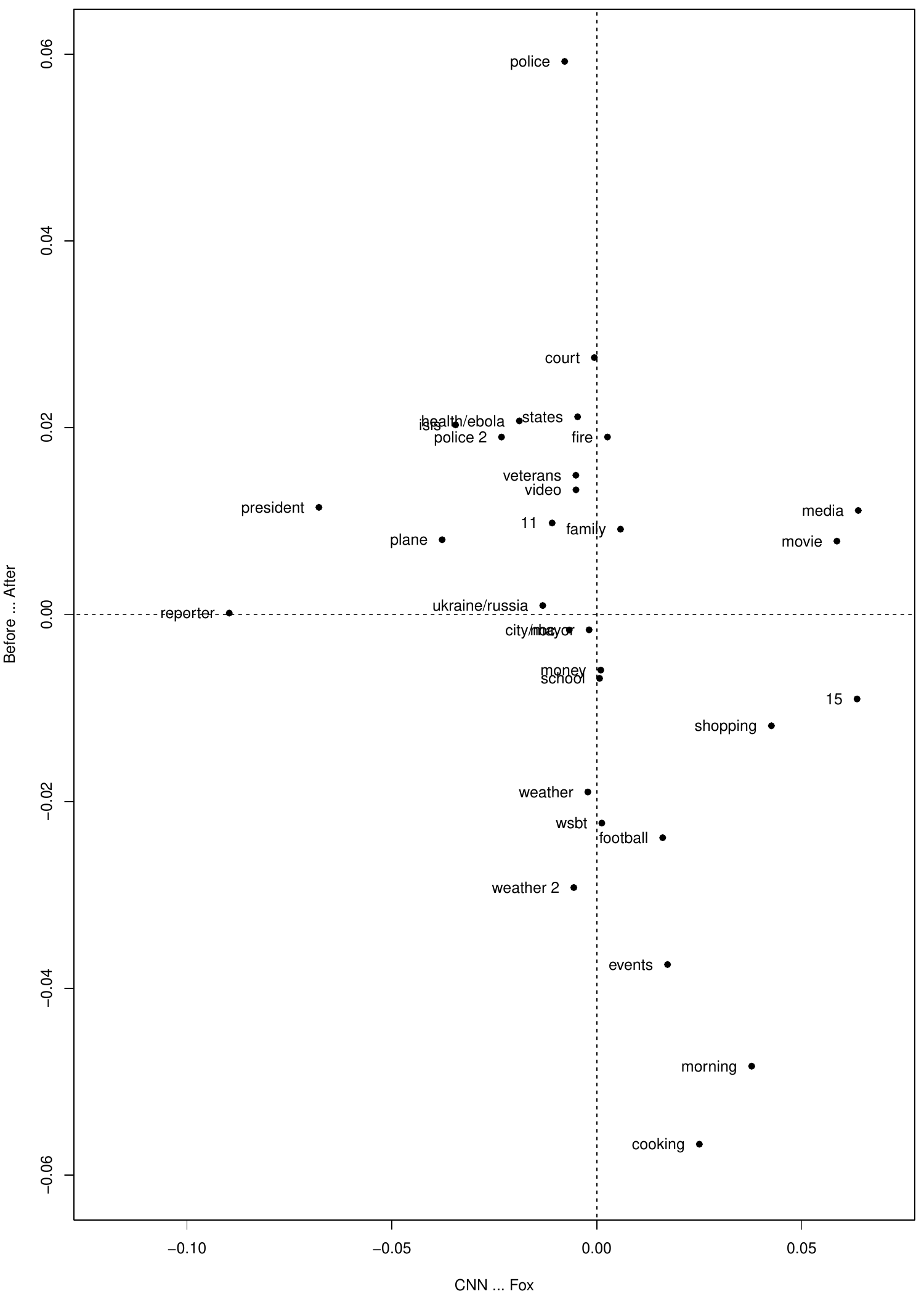}
    \caption{Results for STM on \textbf{2014 data}. Change in topic proportion shifting from CNN to Fox on the x-axis and before Sinclair purchase to after Sinclair purchase on the y-axis.}
    \label{fig:stm-2014-appendix}
\end{figure*}

\newpage 
\begin{figure*}
    \centering
    \includegraphics[width=0.9\linewidth]{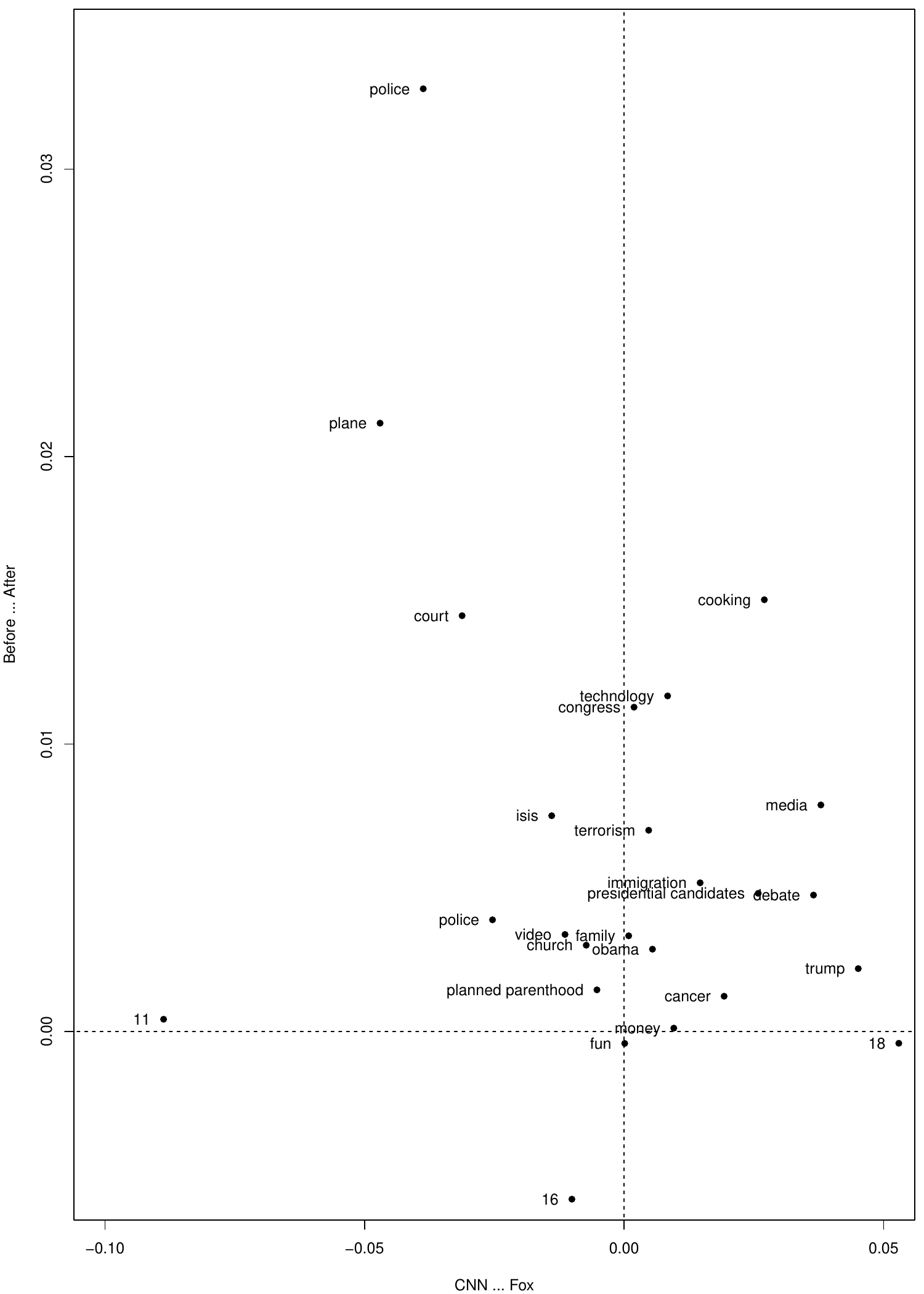}
    \caption{Results for STM on \textbf{2015 data}. Change in topic proportion shifting from CNN to Fox on the x-axis and before Sinclair purchase to after Sinclair purchase on the y-axis. Topics 6 (animals), 8 (station), 12 (weather), 14 (football), and 20 (school) are omitted, but can be found in Figure \ref{fig:stm-2015-main}.}
    \label{fig:stm-2015-appendix}
\end{figure*}

\newpage 
\begin{figure*}
    \centering
    \includegraphics[width=0.9\linewidth]{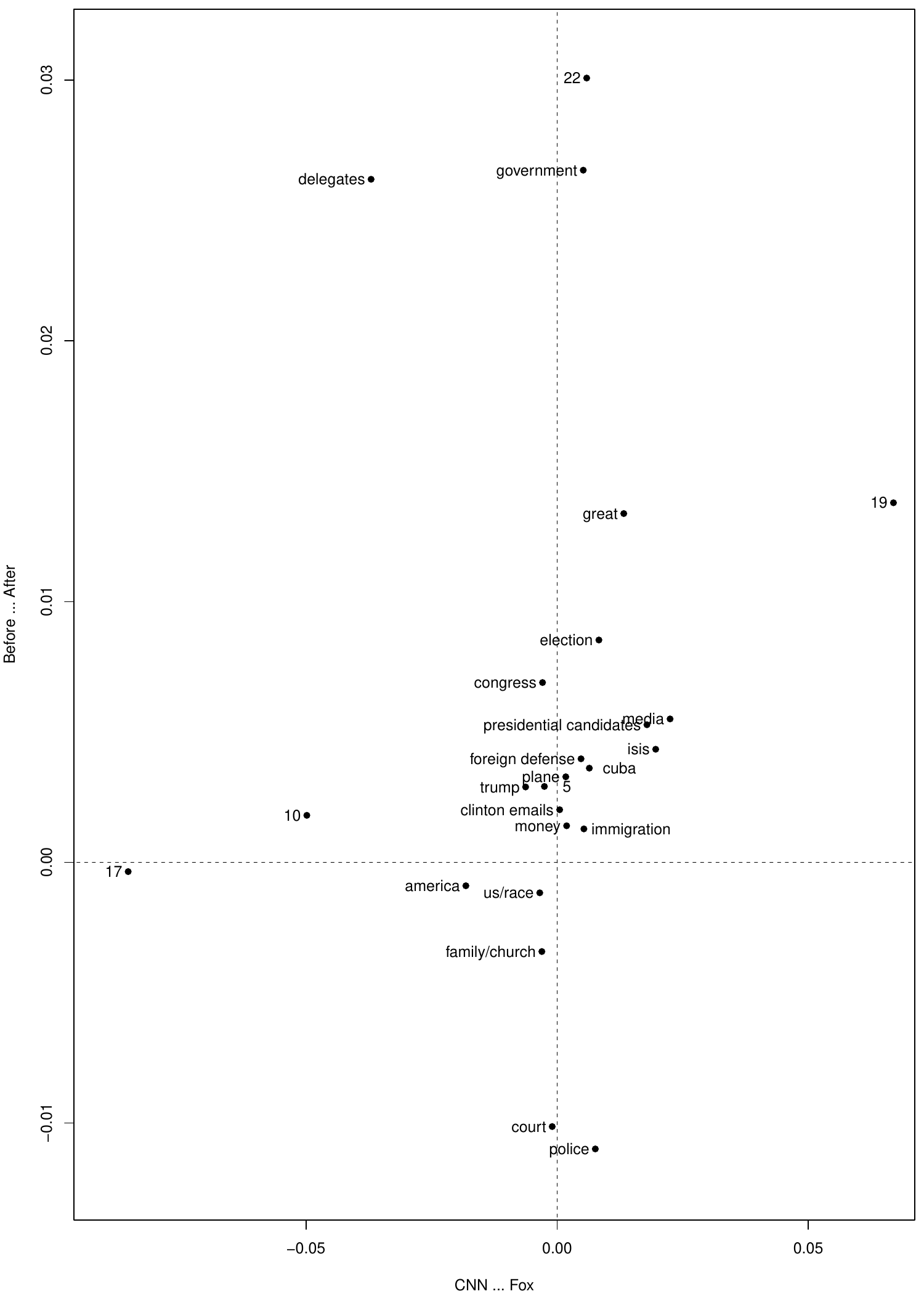}
    \caption{Results for STM on \textbf{2016 data}. Change in topic proportion shifting from CNN to Fox on the x-axis and before Sinclair purchase to after Sinclair purchase on the y-axis. Topics 4 (family), 9 (football), 11 (weather), 24 (terrorism), and 27 (health) are omitted, but can be found in Figure \ref{fig:stm-2016-main}.}
    \label{fig:stm-2016-appendix}
\end{figure*}

\begin{longtable}{c|p{12cm}}

\caption{Topics for the topical content STM for Sinclair purchased station data.}\label{tab:content-topics}
\\
\hline 
\textbf{Topic} & \textbf{Words} \\
\hline
\endfirsthead
\hline
\textbf{Topic} & \textbf{Words} \\
\hline
\endhead
1 & cameras, photos, captured, photo, wjla, footage, images \\
2 & climb, time \\
3 & dancing, parade, dance, candy, cheer, band, golf \\
4 & videos, video, posted, media, facebook, twitter, shows \\
5 & honored, behalf, introduce, celebrate, colleagues, supportive, achieve \\
6 & commissioners, proposal, properties, council, citys, apartments, mayor \\
7 & emergency, drug, staffing, deaths, drugs, procedures, prevention \\
8 & bristol, friendship, flag, ford, williams, courage, flags \\
9 & touchdowns, touchdown, undefeated, quarterback, halftime, scored, coach \\
10 & spicy, flour, yummy, chocolate, garlic, recipes, flavors \\
11 & violence, americans, weapons, global, domestic, religious, border \\
12 & tea, super, ideas, magic, style, interesting, traditional \\
13 & flames, firefighters, firefighter, crash, smoke, crashed, blaze \\
14 & flooding, snowfall, flooded, sidewalks, snow, pipe, water \\
15 & joy, excited, winners, join, sleeping, wave, scary \\
16 & artists, donations, museum, concert, donate, festival, organizers \\
17 & animals, animal, humane, dogs, dog, adoption, pet \\
18 & prosecutors, sentenced, prosecutor, courtroom, pleaded, convicted, lawsuit \\
19 & bridge, drivers, lanes, bridges, engineers, gas, crashes \\
20 & thunderstorms, showers, inland, gusts, clouds, sunshine, winds \\
21 & redskins, players, baseball, nfl, playoff, football, player \\
22 & greg, farmers, tropical, sunny, corn, fishing, clay \\
23 & teachers, teacher, students, superintendent, academic, elementary, colleges \\
24 & book, guy, dad, god, married, cuz, mom \\
25 & gunshot, gunman, police, suspects, suspect, custody, homicide \\
26 & workforce, governments, companies, contractors, infrastructure, consumers, taxpayers \\
27 & lawmakers, republicans, senate, congress, democrats, bipartisan, legislature \\
28 & winnebago, sturgeon, marinette, dnr, deer, lakes, oshkosh \\
29 & roz, plater, year-old, rockville, cheetah, -year-old, grandmother \\
30 & sort, obviously, folks, interesting, mentioned, necessarily, maybe \\
\hline
\end{longtable}

\rowcolors{1}{}{} 

\begin{longtable}{c|c|p{11cm}}

\caption{Topic-Covariate Interactions for the topical content STM for Sinclair purchased station data.}\label{tab:content-topics-beforeafter}
\\
\hline
\textbf{Topic} & \textbf{Group} & \textbf{Words} \\
\hline
\endfirsthead
\hline
\textbf{Topic} & \textbf{Group} & \textbf{Words} \\
\hline
\endhead

\multirow{2}{*}{1} & Before & girl, ray, hurt, hadnt, pictures, capture, submit \\
                   & After  & cell, body, surveillance, recorded, light, footage, speed \\
\hline
\multirow{2}{*}{2} & Before & climb \\
                   & After  & - \\
\hline
\multirow{2}{*}{3} & Before & katie, carrie, congratulations, entertainment, audience, anniversary, stage \\
                   & After  & shoppers, gifts, gift, toys, sale, shopping, eve \\
\hline
\multirow{2}{*}{4} & Before & post, page, youtube, facebook, twitter, recognize, shows \\
                   & After  & social, moment, pictures, released, heres, online, moments \\
\hline
\multirow{2}{*}{5} & Before & brad, wedding, bell, lisa, birthday, hair, awards \\
                   & After  & rhode, island, providence, lord, assembly, pray, governor \\
\hline
\multirow{2}{*}{6} & Before & annie, copeland, roth, rick, bend, wsbts, spencer \\
                   & After  & taxpayers, revenue, funding, taxpayer, taxes, bern, funds \\
\hline
\multirow{2}{*}{7} & Before & security, cyber, consumer, digital, cell, sheriffs, sheriff \\
                   & After  & symptoms, hospitals, virus, patients, disease, cancer, goshen \\
\hline
\multirow{2}{*}{8} & Before & veterans, afghanistan, cemetery, military, war, soldiers, warwick \\
                   & After  & virginia, heather, marion, bank, loan, johnny, richmond \\
\hline
\multirow{2}{*}{9} & Before & irish, tournament, penn, hockey, carl, benton, kimberly \\
                   & After  & greenville, friendship, vegetable, henry, kingsport, science, suv \\
\hline
\multirow{2}{*}{10} & Before & festival, easter, holiday, book, babies, amanda, reduce \\
                    & After  & pauline, classroom, flowers, meteorologist, birds, awesome, fitness \\
\hline
\multirow{2}{*}{11} & Before & cancer, patients, disease, therapy, symptoms, diagnosed, shutdown \\
                    & After  & afghanistan, russia, terrorist, iraq, terror, troops, soldiers \\
\hline
\multirow{2}{*}{12} & Before & emily, pauline, deem, angela, colors, rachel, cute \\
                    & After  & gop, donald, trump, cruz, hillary, democratic, republican \\
\hline
\multirow{2}{*}{13} & Before & tornado, suv, homeowner, korff, lightning, suspicious, insurance \\
                    & After  & jeanette, reyes, trains, rail, metro, flights, riders \\
\hline
\multirow{2}{*}{14} & Before & cherry, heat, ski, storms, supply, residential, restrictions \\
                    & After  & boats, beaches, roadway, bern, coastal, intersection, atlantic \\
\hline
\multirow{2}{*}{15} & Before & dinner, excited, winners, scary, sleeping, joy, wave \\
                    & After  & joining, wave, joy, excited, winners, sleeping, join \\
\hline
\multirow{2}{*}{16} & Before & shopping, shoppers, christmas, sales, gift, gifts, volunteer \\
                    & After  & ceremony, tribute, parade, exhibit, veterans, breast, memorial \\
\hline
\multirow{2}{*}{17} & Before & egg, eggs, breakfast, forecast, population, deer, sleeping \\
                    & After  & racing, farm, lisa, race, girl, races, phil \\
\hline
\multirow{2}{*}{18} & Before & arrests, suspected, recorded, ford, steven, dcs, cranston \\
                    & After  & murder, homicide, fairfax, murdered, custody, baltimore, death \\
\hline
\multirow{2}{*}{19} & Before & trains, rail, riders, passengers, metro, intersection, airport \\
                    & After  & consumer, consumers, fees, cents, revenue, dollars, costs \\
\hline
\multirow{2}{*}{20} & Before & snow, slippery, marinette, ecu, oshkosh, appleton, pete \\
                    & After  & macon, debris, warner, interstate, wgxa, trees, houston \\
\hline
\multirow{2}{*}{21} & Before & marathon, runners, race, receiver, races, racing, kickoff \\
                    & After  & hockey, tournament, jordan, championship, basketball, coaching, teammates \\
\hline
\multirow{2}{*}{22} & Before & garden, flowers, boat, soil, farm, boats, lawn \\
                    & After  & amanda, elkhart, deputies, sheriffs, wsbt, bend, wgxa \\
\hline
\multirow{2}{*}{23} & Before & toys, technology, santa, talented, pilot, projects, connecticut \\
                    & After  & athlete, athletes, craven, buses, trauma, pandemic, ecu \\
\hline
\multirow{2}{*}{24} & Before & theater, movie, inspired, baby, mothers, wsb, library \\
                    & After  & guys, robert, egg, sport, skill, congratulations, retirement \\
\hline
\multirow{2}{*}{25} & Before & surveillance, rousey, metro, accused, morgan, beaten, cell \\
                    & After  & detective, protesters, autopsy, murder, protest, trauma, departments \\
\hline
\multirow{2}{*}{26} & Before & taxes, prices, tax, bills, sales, fees, cents \\
                    & After  & cyber, intelligence, pentagon, privacy, homeland, matters, providers \\
\hline
\multirow{2}{*}{27} & Before & mayor, mayors, vincent, candidates, sector, clinton, brianne \\
                    & After  & immigration, tax, shutdown, proposal, marijuana, funding, proposed \\
\hline
\multirow{2}{*}{28} & Before & doran, beth, schlicht, follow-, alex, ship, chad \\
                    & After  & montana, kalispell, bozeman, montanas, missoula, butte, snow \\
\hline
\multirow{2}{*}{29} & Before & homicide, detective, body, cruz, eve, lord, reyes \\
                    & After  & rousey, brianne, patrice, graham, warwick, cemetery, girl \\
\hline
\multirow{2}{*}{30} & Before & realize, waited, worried, biggest, opened, days, deals \\
                    & After  & bruce, yeah, conversations, convention, buildings, appreciate, helpful \\
\hline
\end{longtable}

\begin{figure*}[htbp]
    \centering
    \begin{subfigure}[b]{0.45\linewidth}
        \centering
        \includegraphics[width=\linewidth]{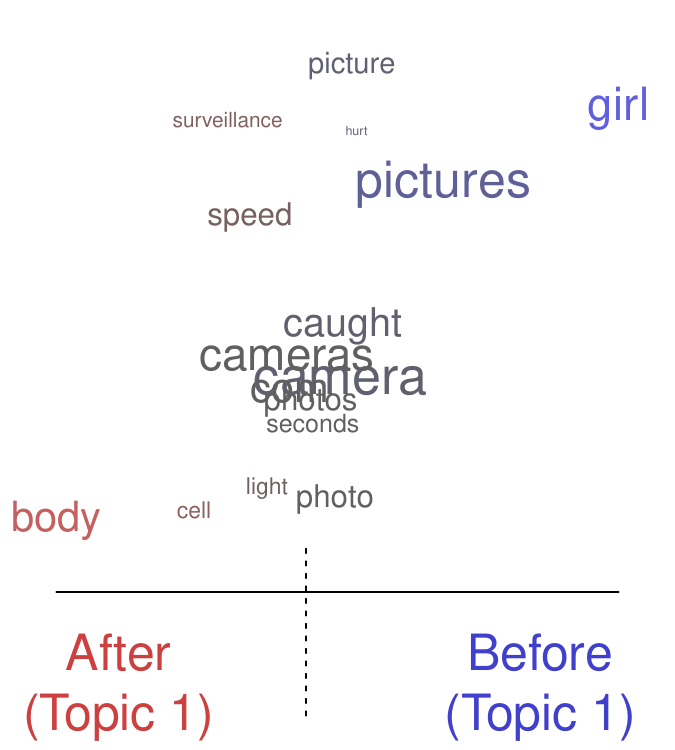}
        \caption{}
        \label{fig:content-topic1}
    \end{subfigure}
    \hfill
    \begin{subfigure}[b]{0.45\linewidth}
        \centering
        \includegraphics[width=\linewidth]{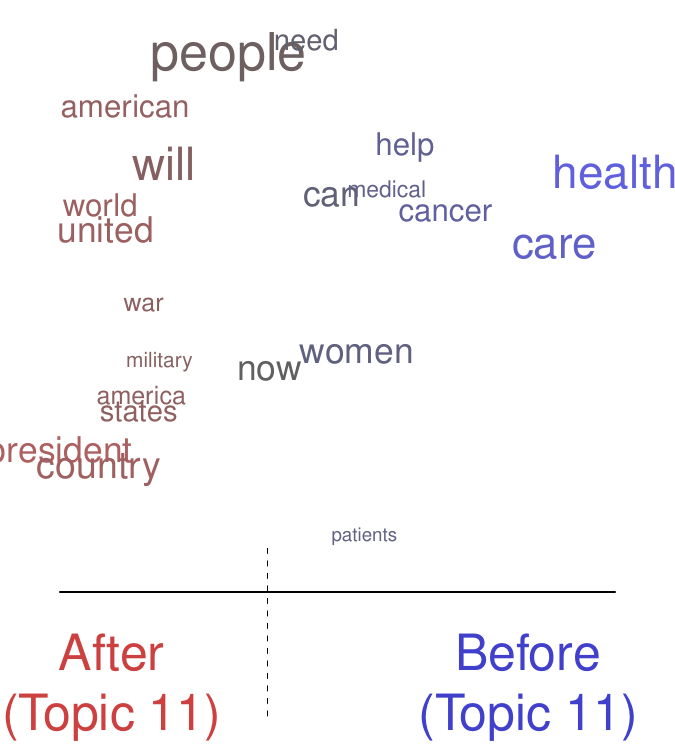}
        \caption{}
        \label{fig:content-topic11}
    \end{subfigure}
    
    \vspace{0.5cm}
    
    \begin{subfigure}[b]{0.45\linewidth}
        \centering
        \includegraphics[width=\linewidth]{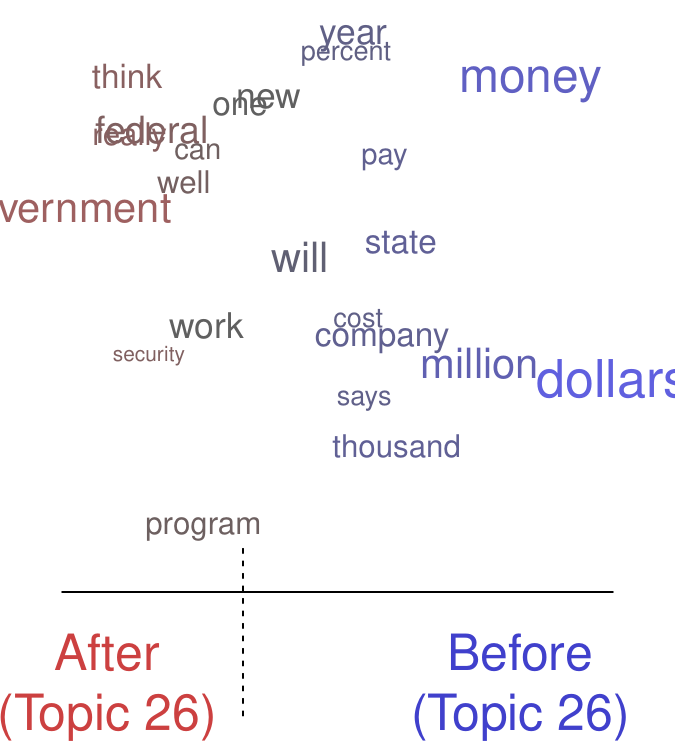}
        \caption{}
        \label{fig:content-topic26}
    \end{subfigure}
    \hfill
    \begin{subfigure}[b]{0.45\linewidth}
        \centering
        \includegraphics[width=\linewidth]{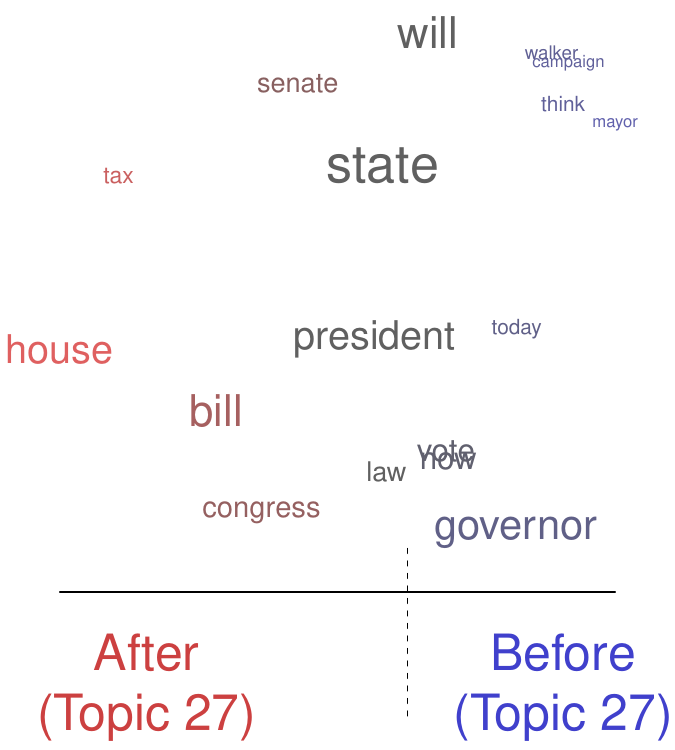}
        \caption{}
        \label{fig:content-topic27}
    \end{subfigure}
    
    \caption{STM with before/after purchase as a topical content covariate. These plots show words within a topic which are strongly associated with coverage before Sinclair takeover as opposed to after Sinclair purchase.}
    \label{fig:content-stm-words}
\end{figure*}

\newpage 

\newpage 
\begin{figure}
    \centering
    \includegraphics[width=0.9\linewidth]{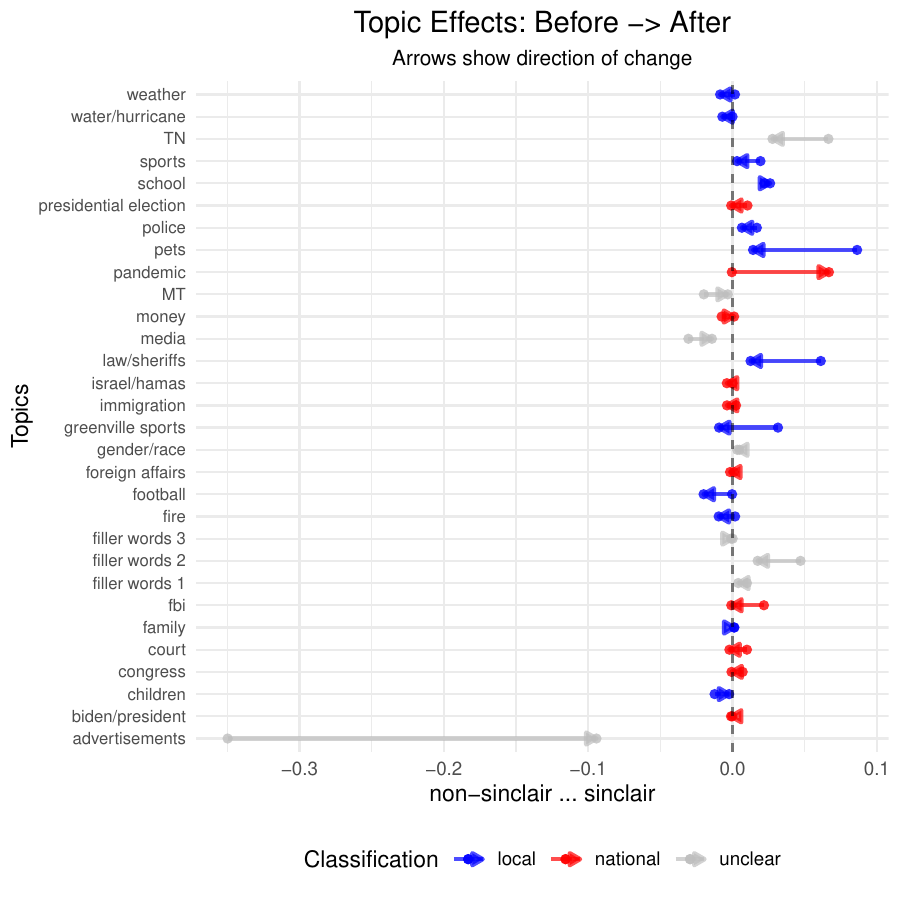}
    \caption{Results for STM on \textbf{all paired data}. Change in topic proportion shifting from non-Sinclair to Sinclair affiliate on the x-axis, and shift before and after purchase date is shown with arrows. Red denotes national topics, blue denotes local topics, and gray topics are unclear. Topic list is shown on the y-axis. The graph with only national/local topics can be found in Figure \ref{fig:paired-all-main}.}
    \label{fig:paired-all-appendix}
\end{figure}

\newpage 
\begin{figure}
    \centering
    \includegraphics[width=0.9\linewidth]{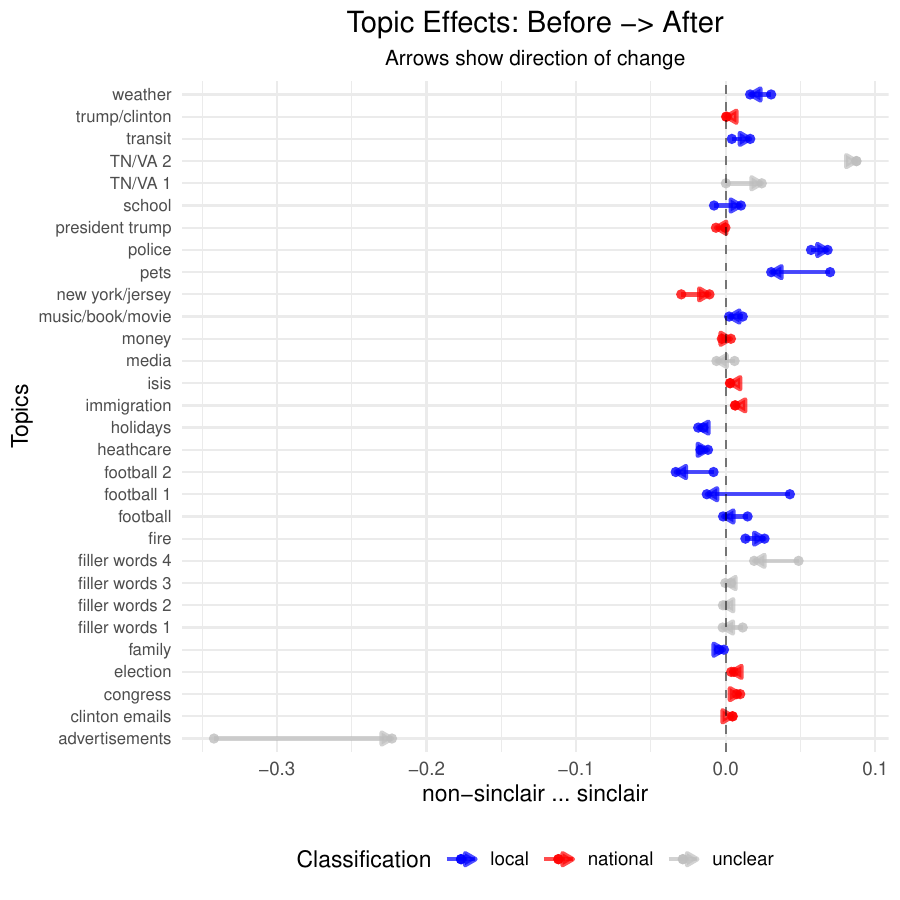}
    \caption{Results for STM on \textbf{paired data before 2020}. Change in topic proportion shifting from non-Sinclair to Sinclair affiliate on the x-axis, and shift before and after purchase date is shown with arrows. Red denotes national topics, blue denotes local topics, and gray topics are unclear. Topic list is shown on the y-axis. The graph with only national/local topics can be found in Figure \ref{fig:paired-pre2020-main}.}
    \label{fig:paired-pre2020-appendix}
\end{figure}

\renewcommand{\arraystretch}{1.2}
\rowcolors{2}{gray!10}{white}
\begin{center}

\begin{longtable}{>{\bfseries}c | >{\bfseries}m{0.08\textwidth} | m{0.15\textwidth} | m{0.15\textwidth} | m{0.15\textwidth} | m{0.15\textwidth}}

\caption{Topics for Paired Analysis STM with all data. See appendix for column information.}\label{tab:topics-paired-all} \\

\hline
\textbf{Topic} & \textbf{Label} & \textbf{Highest Prob} & \textbf{FREX} & \textbf{Lift} & \textbf{Score} \\
\hline
\endfirsthead

\hline
\textbf{Topic} & \textbf{Label} & \textbf{Highest Prob} & \textbf{FREX} & \textbf{Lift} & \textbf{Score} \\
\hline
\endhead

1 & immigra- tion & border, people, country, new, law, governor, city & border, illegal, texas, immigration, migrants, immigrants, cities & asylum, climb, cartels, migrant, migrants, sanctuary, aliens & border, climb, immigration, migrants, immigrants, illegal, sanctuary \\
2 & police & police, car, morning, say, officers, happened, just & scene, police, officers, crash, officer, shooting, vehicle & submit, crash, shooter, scene, injuries, accident, fatal & submit, police, officers, scene, crash, car, injuries \\
3 & court & case, court, trump, judge, will, president, former & supreme, judge, trial, indictment, court, legal, lawyers & testifying, fani, indictment, willis, wade, lawyers, supreme & trump, testifying, court, judge, trial, supreme, donald \\
4 & media & media, news, new, video, fox, york, show & video, media, twitter, fox, tucker, post, movie & video, instagram, twitter, outlets, tucker, platforms, magazine & video, media, fox, twitter, york, social, facebook \\
5 & football & first, going, get, got, game, right, hes & ball, yards, coach, game, gonna, rivals, play & videos, rivals, clock, bennett, pals, snap, powered & videos, touchdown, yards, game, coach, ball, quarterback \\
6 & israel/ hamas & ukraine, israel, war, now, military, will, hamas & hamas, gaza, israel, israeli, forces, hostages, ukrainian & counteroffensive, hamas, hezbollah, hostages, idf, palestinian, casualties & hamas, ukraine, gaza, israel, casualties, russia, putin \\
7 & foreign affairs & president, united, will, states, administration, secretary, foreign & foreign, china, secretary, nuclear, countries, administration, president-elect & foreign, sanctions, korea, nuclear, cuba, taiwan, ambassador & foreign, putin, president, russia, china, nuclear, president-elect \\
8 & law/ sheriffs & county, found, office, law, charges, case, sheriffs & sheriffs, murder, sheriff, arrested, charges, charged, prison & colonel, sheriffs, deputies, homicide, sheriff, arrested, murder & colonel, sheriffs, county, charges, murder, police, investigators \\
9 & TN & city, county, will, says, johnson, tennessee, news & channel, johnson, tennessee, josh, city, kingsport, sarah & champion, newschannel, commissioners, eleven, tennessees, channel, improvements & champion, county, tennessee, kingsport, city, bristol, johnson \\
10 & money & money, dollars, million, tax, pay, jobs, will & tax, jobs, taxes, dollars, companies, money, billion & unbelievable, tax, wages, income, paycheck, investments, medicare & unbelievable, tax, dollars, inflation, taxes, economy, money \\
11 & gender/ race & people, country, women, america, american, will, black & rights, america, black, freedom, hate, women, racist & pledge, religious, racism, protest, freedoms, gender, religion & pledge, america, women, americans, thank, democracy, american \\
12 & water/ hurricane & water, now, area, just, can, right, will & water, bridge, plane, storm, flight, airport, coast & yep, ocean, flight, hurricane, bridge, debris, rail & yep, water, storm, hurricane, river, airport, aircraft \\
13 & advertis- ements & new, christmas, store, now, get, car, bristol & christmas, store, holiday, sale, restaurant, shop, sales & beats, christmas, santa, sale, holiday, stores, restaurant & beats, bristol, christmas, kingsport, furniture, abington, customers \\
14 & congress & house, republicans, bill, republican, senate, democrats, party & senate, speaker, republicans, mccarthy, congressman, capitol, republican & maga, speaker, mccarthy, senate, schumer, senators, mcconnell & maga, republicans, democrats, republican, senate, speaker, congress \\
15 & filler words 1 & way, make, can, get, see, back, sure & way, make, done, long, sure, ways, making & way, shannon, ways, connecting, shape, figure, apart & way, make, can, see, get, done, ways \\
16 & greenville sports & one, tonight, game, green, win, greenville, first & greenville, green, score, devils, daniel, win, blue & chevrolet, devils, warriors, greenville, boone, daniel, finds & chevrolet, greenville, touchdown, devils, game, boone, score \\
17 & filler words 2 & know, think, going, dont, thats, people, like & think, know, mean, dont, youre, thing, kind & flash, mean, sort, know, think, honestly, dont & think, know, flash, mean, going, people, dont \\
18 & pandemic & health, will, cases, state, can, people, county & testing, virus, health, cases, distancing, masks, tests & trusted, virus, vaccinated, distancing, quarantine, testing, outbreak & trusted, health, missoula, virus, kovat, testing, county \\
19 & weather & morning, snow, see, will, weather, going, day & snow, temperatures, showers, weather, rain, forecast, degrees & snow, toss, showers, sunshine, cooler, cloudy, temperatures & toss, snow, temperatures, montana, showers, missoula, kalispell \\
20 & fbi & information, investigation, fbi, department, report, evidence, questions & fbi, letter, classified, chairman, investigation, committee, evidence & heal, fbi, oversight, server, classified, document, letter & fbi, heal, investigation, classified, documents, evidence, committee \\
21 & filler words 3 & gtgt, reporter, said, say, dont, women, gtgtgt & gtgt, reporter, gtgtgt, e-mails, usa, cnn, e-mail & gtgt, gtgtgt, usa, reporter, e-mail, e-mails, aides & gtgt, usa, reporter, gtgtgt, e-mails, e-mail, cnn \\
22 & children & can, children, kids, care, help, parents, child & cancer, children, mental, parents, doctor, child, doctors & tent, cancer, parent, doctors, diagnosed, doctor, pregnant & tent, children, kids, parents, patients, child, hospital \\
23 & family & just, family, years, like, life, time, know & book, friends, family, life, mom, loved, father & jay, lord, funeral, grandfather, queen, larry, mom & jay, family, book, father, life, mom, thank \\
24 & fire & fire, smoke, firefighters, burning, fires, tha, burn & tha, thi, tth, firefighters, ths, whe, fire & ant, ere, tha, tht, tit, ahe, aim & fire, aim, tth, tha, thi, firefighters, ath \\
25 & presiden- tial election & trump, donald, clinton, campaign, hillary, election, hes & hillary, clinton, donald, debate, trump, campaign, voters & gentlemen, hampshire, romney, rnc, hillary, mitt, battleground & trump, donald, clinton, hillary, gentlemen, election, voters \\
26 & pets & right, just, little, got, can, like, yeah & dog, little, yeah, beautiful, fun, love, okay & tail, adoption, dogs, dog, chicken, delicious, sugar & tail, gonna, dog, adoption, yeah, thank, fun \\
27 & MT & montana, says, missoula, people, help, community, will & mtn, montanas, park, helena, medicine, missoula, montana & medicine, nbcmontanacom, wildlife, mtns, helena, mtn, recreation & medicine, montana, missoula, mtn, flathead, bozeman, montanas \\
28 & school & school, students, community, year, just, schools, kids & students, music, school, campus, schools, student, university & music, arts, festival, campus, teachers, students, classes & music, students, school, campus, schools, kids, community \\
29 & biden/ president & biden, joe, president, dont, will, hunter, people & biden, joe, hunter, bidens, greg, brian, jesse & anti, jeanine, kamala, ainsley, bidens, carley, newsom & biden, bidens, joe, anti, hunter, president, democrats \\
30 & sports & game, team, play, just, playing, know, got & playing, sports, football, team, play, game, field & playing, nfl, soccer, baseball, basketball, sport, league & playing, game, football, montana, coach, sports, players \\
\hline
\end{longtable}

\end{center}

\renewcommand{\arraystretch}{1.2}
\rowcolors{2}{gray!10}{white}
\begin{center}

\begin{longtable}{>{\bfseries}c | >{\bfseries}m{0.08\textwidth} | m{0.15\textwidth} | m{0.15\textwidth} | m{0.15\textwidth} | m{0.15\textwidth}}

\caption{Topics for Paired Analysis STM with data before 2020. See appendix for column information.}\label{tab:topics-paired-pre2020} \\

\hline
\textbf{Topic} & \textbf{Label} & \textbf{Highest Prob} & \textbf{FREX} & \textbf{Lift} & \textbf{Score} \\
\hline
\endfirsthead

\hline
\textbf{Topic} & \textbf{Label} & \textbf{Highest Prob} & \textbf{FREX} & \textbf{Lift} & \textbf{Score} \\
\hline
\endhead

1 & new york/ jersey & new, york, national, times, record, jersey, join & york, new, jersey, national, record, join, miller & jooy, york, jersey, new, miller, join, record & new, york, jooy, national, jersey, join, record \\
2 & media & women, media, news, said, press, fox, saying & women, media, twitter, press, sexual, comments, magazine & journalists, sexually, ypcom, fzone, twitter, women, magazine & women, media, sexual, ypcom, fox, twitter, journalists \\
3 & football 1 & one, tonight, first, game, win, nothing, back & score, quarter, final, nothing, devils, lady, touchdown & aint, warriors, daniel, gate, scores, boone, hits & aint, touchdown, devils, score, greenville, game, crockett \\
4 & filler words 1 & can, way, make, get, want, thats, sure & way, make, can, sure, lets, put, done & way, ways, make, sure, can, try, glad & way, can, lets, make, want, talk, sure \\
5 & president trump & trump, hes, donald, president, president-elect, secretary, trumps & president-elect, transition, cabinet, romney, mitt, trumps, secretary & upload, cabinet, mattis, flynn, giuliani, sessions, bolton & trump, president-elect, donald, romney, secretary, trumps, mitt \\
6 & money & jobs, money, going, business, tax, will, million & jobs, tax, companies, billion, economy, money, obamacare & unbelievable, trillion, jobs, billion, companies, carrier, regulations & unbelievable, tax, jobs, obamacare, taxes, economy, companies \\
7 & filler words 2 & gtgt, reporter, say, dont, gtgtgt, said, cnn & gtgt, reporter, gtgtgt, cnn, jake, videos, wolf & videos, gtgt, gtgtgt, reporter, cnns, cnn, jake & gtgt, videos, reporter, gtgtgt, cnn, cnns, jake \\
8 & holidays & great, got, right, well, just, christmas, come & christmas, fun, parade, event, excited, tickets, folks & palswebcom, merry, christmas, parade, festival, santa, celebration & christmas, fun, parade, palswebcom, festival, daytime, merry \\
9 & clinton emails & clinton, fbi, information, investigation, emails, election, hillary & fbi, emails, e-mails, comey, email, classified, cyber & classified, heal, hacked, hackers, e-mails, hack, hacking & clinton, fbi, e-mails, heal, hillary, comey, investigation \\
10 & immigr- ation & people, country, law, will, president, americans, america & immigration, rights, immigrants, flag, illegal, americans, law & immigrants, pledge, sanctuary, religion, protests, religious, constitution & pledge, immigration, immigrants, sanctuary, law, americans, federal \\
11 & TN/VA 1 & city, johnson, will, bristol, kingsport, street, now & johnson, city, bristol, kingsport, project, downtown, champion & champion, construction, citys, johnson, city, project, bristol & champion, johnson, city, kingsport, bristol, downtown, tri-cities \\
12 & filler words 3 & john, space, don, bob, hero, beer, wine & ray, hero, tth, don, tin, ship, bob & ath, aan, ahe, tha, whe, aon, ihe & ray, tth, ihe, tnd, ahe, tin, tng \\
13 & music/ book/ movie & like, know, one, really, yeah, show, just & book, music, movie, film, song, show, love & music, movie, movies, book, film, songs, sing & music, movie, book, film, song, yeah, love \\
14 & weather & morning, today, now, day, will, see, well & morning, tomorrow, weekend, afternoon, sunday, hours, live & webcom, temperatures, morning, forecast, tomorrow, wednesday, rain & morning, webcom, tomorrow, weather, temperatures, rain, forecast \\
15 & health- care & health, care, can, hospital, medical, help, also & health, patients, cancer, medical, disease, doctors, treatment & doctors, trusted, medication, symptoms, disease, patients, cancer & health, patients, trusted, hospital, medical, disease, doctors \\
16 & isis & isis, war, president, military, will, united, now & isis, syria, iran, nuclear, military, iraq, forces & mosul, pit, sanctions, assad, iranian, iraqi, nato & isis, syria, russia, iran, pit, iraq, putin \\
17 & football & team, game, playing, play, season, year, football & playing, players, games, sports, basketball, football, team & playing, nfl, league, basketball, players, baseball, athletes & playing, game, football, players, coach, games, etsu \\
18 & TN/VA 2 & county, says, tennessee, state, will, news, virginia & county, josh, tennessee, sarah, carter, board, sullivan & sponsored, defuse, jackie, burnie, commissioner, nate, tennessees & county, tennessee, sponsored, sullivan, unicoi, channel, defuse \\
19 & fire & fire, now, people, one, officials, just, attack & fire, scene, alert, attack, firefighters, authorities, montana & update, firefighters, fire, shooter, fires, alert, flames & update, fire, firefighters, police, officials, montana, authorities \\
20 & police & police, said, case, say, officers, officer, man & officer, sheriffs, officers, murder, police, charges, charged & year-old, submit, deputies, sheriffs, murder, jury, aggravated & police, submit, sheriffs, officers, investigators, investigation, charges \\
21 & football 2 & gonna, first, now, back, get, game, hes & ball, gonna, science, thomas, yards, yard, touchdown & chevrolet, thomas, clock, bennett, rivals, crockett, ball & chevrolet, crockett, touchdown, yards, gonna, game, ball \\
22 & transit & car, water, just, road, get, plane, train & water, miles, crash, plane, train, bus, car & slow, flight, miles, crashed, drivers, engine, driver & slow, crash, water, car, highway, driver, plane \\
23 & filler words 4 & know, think, going, people, dont, well, like & think, know, mean, dont, going, youre, thing & aim, mean, tucker, sort, know, think, neil & think, know, mean, going, people, aim, dont \\
24 & congress & president, house, republican, party, obama, democrats, republicans & senate, senator, republicans, democrats, republican, cruz, governor & pelosi, ron, rubio, senate, jeb, christie, cruz & republican, democrats, obama, republicans, ron, senate, president \\
25 & pets & little, right, just, like, look, got, yeah & dog, adoption, animals, shelter, dogs, animal, okay & animals, appalachian, adoption, adorable, chocolate, kitchen, cream & appalachian, adorable, adoption, shelter, gonna, dog, animal \\
26 & election & trump, clinton, election, vote, donald, hillary, states & voting, vote, electoral, votes, voters, michigan, polls & mexican, electoral, battleground, recount, electorate, stein, rigged & trump, clinton, hillary, election, donald, voters, electoral \\
27 & advertis- ements & now, home, store, free, one, get, buy & sale, sales, furniture, store, shop, shopping, buy & wallace, accessories, furniture, sales, app, sale, brands & wallace, furniture, sale, sales, store, kingsport, abington \\
28 & family & family, just, people, life, know, help, years & family, children, families, father, mother, life, mom & properties, funeral, mom, journey, honor, mothers, moms & properties, family, children, veterans, kids, church, mother \\
29 & trump/ clinton & trump, donald, clinton, hillary, debate, think, said & debate, hillary, clinton, donald, trump, shes, candidates & absolute, moderator, debate, debates, lester, temperament, universe & trump, clinton, hillary, donald, debate, absolute, clintons \\
30 & school & school, students, video, schools, university, college, kids & video, students, campus, school, schools, student, elementary & video, campus, elementary, teachers, teacher, classes, students & video, school, students, campus, student, schools, gun \\
\hline
\end{longtable}

\end{center}

\end{document}